%% file: neurips_data_2024.tex
\documentclass{article}

\usepackage[preprint,nonatbib]{neurips_data_2024}

\usepackage[utf8]{inputenc} 
\usepackage[T1]{fontenc}    
\usepackage{hyperref}       
\usepackage{url}            
\usepackage{booktabs}       
\usepackage{amsfonts}       
\usepackage{nicefrac}       
\usepackage{microtype}      

\usepackage{color,array}
\usepackage{graphicx}
\graphicspath{{images/}}
\usepackage{subcaption}
\usepackage{graphicx}
\usepackage{caption}
\usepackage{tabularx,booktabs}
\usepackage[dvipsnames]{xcolor}

\usepackage{algorithm,algorithmic}

\title{Dual Thinking and Logical Processing - Are Multi-modal Large Language Models Closing the Gap with Human Vision ?}

\author{%
  Kailas Dayanandan\\
  Indian Institute of Technology \\
  Delhi, 110016 \\
  \texttt{kailasd@gmail.com} \\
  \And
  Nikhil Kumar\\ 
  IIIT \\
  Delhi, 110020 \\  
  \texttt{nikhil21174@iiitd.ac.in} \\
  \And
  Anand Sinha\\ 
  Indian Institute of Technology \\
  Delhi, 110016 \\  
  \texttt{anand.sinha85@gmail.com} \\
  \And
  Brejesh Lall \\
  Indian Institute of Technology \\
  Delhi, 110016 \\
  \texttt{brejesh@ee.iitd.ac.in} \\
}
\begin{document}

\maketitle

\input{sec/0_abstract}

\input{sec/1_intro}
\input{sec/2_related_work}
\input{sec/3_dual_thinking}

\input{sec/4_methods}
\input{sec/5_results}
\input{sec/7_discussion}

\input{sec/8_conclusion}

\clearpage

\bibliography{references}{}
\bibliographystyle{plain}

\end{document}

%% file: sec/0_abstract.tex
\begin{abstract}
The dual thinking framework considers fast, intuitive, and slower logical processing. The perception of dual thinking in vision requires images where inferences from intuitive and logical processing differ, and the latter is under-explored in current studies. We introduce a novel adversarial dataset to provide evidence for the dual thinking framework in human vision, which also facilitates the study of the qualitative behavior of deep learning models. Our psychophysical studies show the presence of multiple inferences in rapid succession, and analysis of errors shows that the early stopping of visual processing can result in missing relevant information. MLLMs (Multi-modal Large Language Models) and VLMs (Vision Language Models) have made significant progress in correcting errors in intuitive processing in human vision and showed enhanced performance on images requiring logical processing. However, their improvements in logical processing have not kept pace with their advancements in intuitive processing. In contrast, segmentation models exhibit errors similar to those seen in intuitive human processing and lack understanding of sub-structures, as indicated by errors related to sub-components in identified instances. As AI (Artificial Intelligence)-based systems find increasing applications in safety-critical domains like autonomous driving, the integration of logical processing capabilities becomes essential. This not only enhances performance but also addresses the limitations of scaling-based approaches while ensuring robustness and reliability in real-world environments.
\end{abstract}

%% file: sec/1_intro.tex
\section{Introduction}
\label{sec:intro}

Dual thinking framework argues that humans have a fast intuitive system and a slow logical system, and has been widely mentioned in recent years \cite{vanrullen2007power,van2020going,mohsenzadeh2018ultra,daniel2017thinking,chen2021cooperative}. A growing body of electrophysiological research suggests gist formation in initial $150-200ms$ using fast feed-forward processing, followed by slower iterative refinement \cite{vanrullen2007power,grootswagers2019representational,kreiman2020beyond,thorpe1996speed,van2020going,tang2018recurrent}. The dual thinking framework is not part of studies on visual perception due to the unavailability of methods to study them. Recent advancements in state-of-the-art performance have primarily been driven by scaling models and training on massive datasets using extensive computational resources. However, as the performance gains from scaling begin to saturate \cite{kaplan2020scaling}, the focus is shifting toward enhancing the reasoning abilities of foundation models and large language models.

Gestalt principles have been studied for over a century \cite{wagemans2012century,van2023pragnanz} indicating its presence in human visual processing \cite{pomerantz2011grouping}. Gestalt theory proposes that humans perceive things as a whole rather than as a sum of parts \cite{wertheimer1912experimentelle,pomerantz1977perception}, which was a deviation from structuralism that considered analyzing parts as a key to understanding the whole object \cite{wagemans2012century,wagemans2012century2}. Gestalt principles and Structuralism are prominent theories about human visual perception \cite{biscione2023mixed,kim2021neural}; however, they are insufficient to describe class identification and complex logical processing in human vision. The theories are also independent of findings about initial gist formation in $200ms$ observed in many electrophysiological studies \cite{vanrullen2007power,kreiman2020beyond,mohsenzadeh2018ultra,grootswagers2019representational}. There exists a research gap for a dataset that enables the analysis of initial gist formation and logical processing in human vision.

Marr’s three-level hypothesis \cite{anderson2015can,love2015algorithmic,peebles2015thirty} provides a framework for analyzing human visual processing. The analysis of computational aspects or the algorithmic component has been ignored in recent years as end-to-end deep learning models gained prominence \cite{zahn1971graph,papari2008adaptive,shen2015gestalt}. The comparative studies have shown to help improve accuracy, robustness, and generalizability of models by incorporating strategies from human vision  \cite{muttenthaler2024improving,muttenthaler2024improving,geirhos2020shortcut,dapello2020simulating,linsleylearning,linsley2020stable,fel2022harmonizing}. The computational models using deep learning helped enhance the understanding of human vision \cite{vogelsang2018potential,kreiman2020beyond,saxe2021if,kim2019disentangling,lake2017building}, though, they face criticism for using classification models unlike human vision that localizes objects \cite{bowers2022deep,serre2019deep,biscione2023mixed}. Our main contributions in this paper include

\textbf{(a)} We observe that logical processing in the dual thinking framework is essential to define human visual processing completely. Our study shows that logical processing was nascent in segmentation models, and multi-modal LLMs in recent years have made tremendous progress in resolving many logical errors; however, they still face limitations that require reasoning using minor features like size differences in sub-components in an instance.

\textbf{(b)} Our analysis offers insights into intuitive processing and the logical errors addressed in the later stages of human vision. Our study highlights the role of Gestalt principles in sub-component formation and emphasizes the importance of shape in grouping elements during intuitive processing. Additionally, we observed that segmentation models like YOLACT exhibit characteristics similar to intuitive processing in human vision.

\textbf{(c)} We introduce the Human Confusion Dataset, a diverse dataset to study dual thinking and strategies in human vision. We leverage the dataset to study the behavioral properties of deep learning models.

%% file: sec/2_related_work.tex
\section{Related Work}
\label{sec:relatedwork}

Deep learning models are attaining human-level performance in many tasks; however, they lack the robustness and generalization of human vision, which led to a comparison of strategies in deep learning models and human vision \cite{baker2018deep,geirhos2018imagenet,hermann2019origins}. The comparisons are based on external behavior  \cite{geirhos2020beyond,geirhos2021partial,geirhos2019generalisation,tuli2021convolutional} or by comparing internal representations \cite{cadieu2014deep,jacob2021qualitative,mehrer2020individual,raghu2021vision}. In the initial part of this section, we examine the dual-thinking framework and explore studies that compare human vision with deep learning models. Subsequently, we discuss existing datasets, highlighting their limitations in capturing the dual-thinking framework in human vision.

\subsection{Dual Thinking} 
The dual thinking process can be modeled to include an initial feed-forward stage, followed by later-stage generative processing, as present in current studies. For example, Tscshantz introduces hybrid predictive coding, which integrates generative and predictive approaches to visual perception \cite{tscshantz2023hybrid}. Similarly, studies employing generative classifiers have demonstrated their alignment with human performance in classification tasks \cite{jainiintriguing}. However, these studies are predominantly quantitative,  offering limited insights into the underlying reasoning processes in human visual perception, and do not align with current progress in multi-modal LLMs \cite{snell2024scaling}. 

\subsection{Visual Processing} 
As deep learning models have increasingly approached human-level performance in classification tasks, researchers have shifted toward qualitative analyses comparing human vision and deep learning models. A significant finding in these studies is the texture bias in deep learning models \cite{baker2018deep,geirhos2018imagenet} and shape bias in human vision. The robust and accurate models had more shape bias, which led to techniques to increase shape bias in deep learning models \cite{geirhos2018imagenet,xu2020robust,li2020shape,ilyas2019adversarial,subramanian2024spatial,sun2021can}. However, models have shown good performance without shape information (e.g. BagNets \cite{brendel2018approximating}), and shape information can be extracted from later layers despite not being used for classification \cite{hermann2019origins}. Subsequent work also shows that image stylization serves as a strong augmentation with shape bias emerging as a byproduct \cite{mummadi2021does,li2020shape}. While shape may not be essential for discriminatory tasks, these findings highlight the need to explore its significance in tasks such as instance segmentation, which involves object localization and is more similar to human vision. The computational analysis of segmentation task in cognitive science represents segmentation as interval graphs \cite{adolfi2022computationalcogsci,adolfi2022computational,friston2021active}, which do not directly extend to image segmentation as instances can be composed of non-adjacent or disconnected segments (e.g., amodal closure in Fig.\ref{fig:human-amodal}), requiring separate analysis.

\begin{figure*}[ht]
    \centering
    \begin{minipage}{0.88\linewidth}
        \centering
        \subfloat[\label{fig:baker-shape-texture}]{\includegraphics[height=2.2cm]{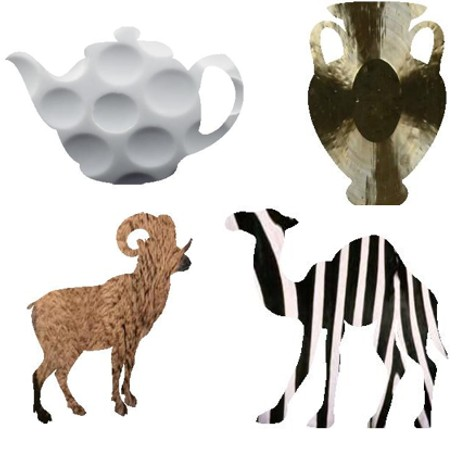}} \hspace{1mm}
        \subfloat[\label{fig:stylized-imagenet-example}]{\includegraphics[height=2.2cm]{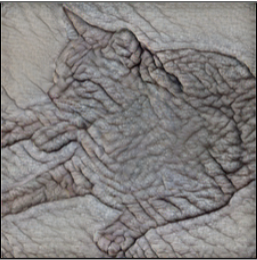}} \hspace{1mm}
        \subfloat[\label{fig:image-abstraction}]{\includegraphics[height=2.2cm]{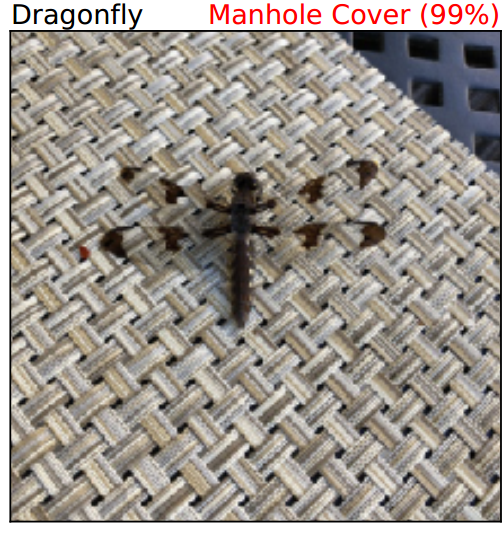}} \hspace{1mm}
        \subfloat[]{\includegraphics[height=2.36cm]{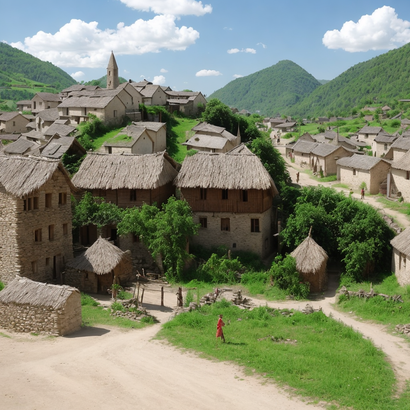}} \hspace{1mm}
        \subfloat[\label{fig:human-hard-vqa}]{\includegraphics[height=2.2cm]{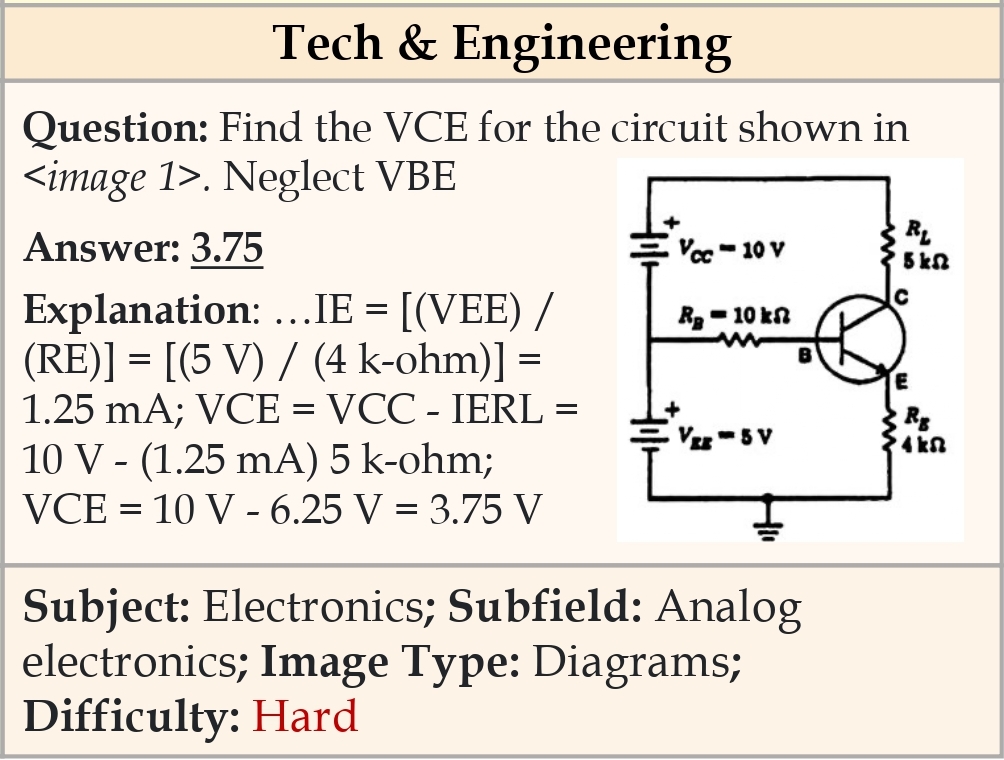}} \hspace{1mm}
        \subfloat[\label{fig:whoops-example}]{\includegraphics[height=2.2cm]{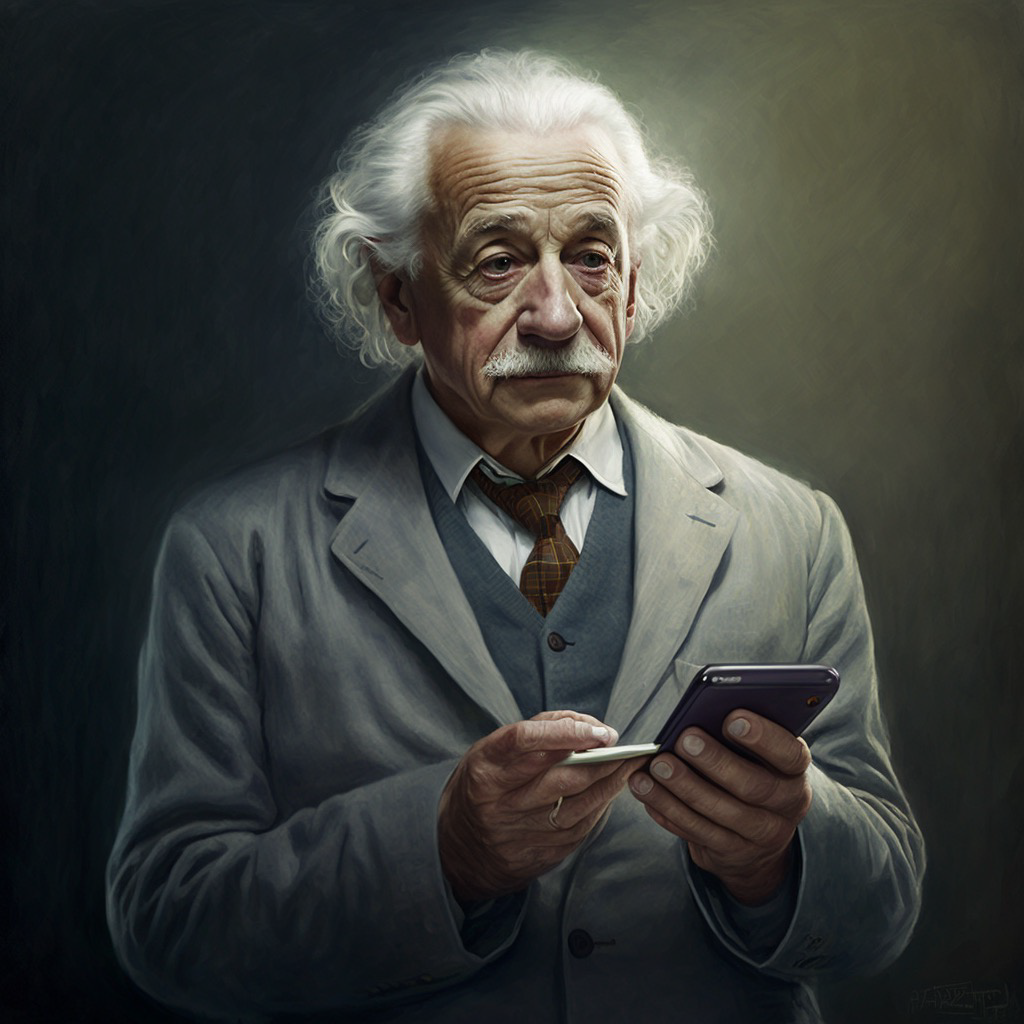}}
    \end{minipage}    
    \caption{        
        (a) Global Shape \cite{baker2018deep} 
        (b) Shape vs Texture from SIN dataset \cite{geirhos2018imagenet} 
        (c) Images in ImageNet-A confuse deep learning models but not human vision \cite{hendrycks2021natural} 
        (d) An example from synthetically generated IllusionBench dataset \cite{hemmat2025hidden} 
        (e) An example from MMMU dataset for MLLMs \cite{yue2024mmmu} 
        (f) Example from WHOOPS dataset showing Einstein using a phone that was not available during his time \cite{bitton2023breaking}.
        }    
\label{fig:VisualTheories}
\end{figure*}

\subsection{Adversarial Datasets} 
Deep learning models made many recent advances, helping them approach human-level performance on many tasks \cite{boutin2023diffusion,boutin2022diversity,jainiintriguing,geirhos2021partial,o2021face,parde2023twin,he2015delving} including on benchmark datasets like ImageNet \cite{russakovsky2015imagenet}. This led to the introduction of many datasets that expose different vulnerabilities of deep learning models \cite{abbas2023progress,alcorn2019strike,hosseini2018semantic,dong2020benchmarking}; including adversarial datasets like Image Net-A, consisting of natural images that were tough for the ResNet-50 \cite{hendrycks2021natural}. However, they are easier for humans and lack instance-level annotations. With deep learning models making progress on many of these adversarial datasets \cite{lee2023robust,li2024survey}, there is a need for datasets that are challenging for human vision to evaluate the progress and understand the behaviors and strategies in human vision and deep learning models. Stylized Image Net (SIN) enforces conflict between texture and shape by synthetically transferring texture from one class to object of another class \cite{geirhos2018imagenet}; however, it lacks diversity as it can study only texture shape bias and does not have a correct ground truth class.
In contrast, images in our dataset have a ground truth class (e.g., Fig. \ref{fig:human-logical-label1} in our dataset is a human hand and not a hand with cat texture, whereas Fig.\ref{fig:stylized-imagenet-example} from SIN is a cat shape with an elephant texture). The images in our datasets have a single stable state compared to bi-stable images. Similarly, another set of datasets focuses only on camouflaged or concealed objects (or figure-ground errors) \cite{fan2020camouflaged,le2021camouflaged,fan2021concealed,lv2021simultaneously,skurowski2018animal}. These datasets cannot evaluate instance segmentation models trained on benchmark MS-COCO dataset \cite{lin2014microsoft} as the class of objects do not match. We do not constrain on any specific cue, which helps us address the gap in research about datasets that help cognitive scientists explore multiple aspects of human visual processing \cite{judd2009learning,linsleylearning,walther2011simple}. Our dataset has images with inherent cue conflicts, with the cue for the correct perception contradicting the cues used for intuitive processing. These images can enable the evaluation of the model qualitatively to understand its inductive biases, which can be helpful while designing systems for use in resource-constrained settings \cite{bachmann2024scaling,xu2021vitae,dayanandan2024enabling}.  

\subsection{Visual Reasoning Datasets} 
Visual question-answering datasets introduced earlier \cite{hudson2019gqa,goyal2017making,marino2019ok,biten2019scene,teney2021unshuffling} are comparatively easy for current large language models. MMMU is a multi-discipline, multi-modal benchmark dataset for visual reasoning with questions on science, technology, and others \cite{yue2024mmmu} and requires specific knowledge of these fields. Similarly, the MATH-Vision dataset has 3,040 mathematical problems with visual contexts sourced from real math competitions \cite{wang2024measuring}. While the images in these datasets have visual information that is simple to understand, answering the questions requires complex analysis and reasoning on the visual information gathered from these images (Fig.\ref{fig:human-hard-vqa}). In contrast, gathering visual information from images in our dataset requires reasoning to select the correct solution from different possible solutions. As deep learning models increasing applications find use in real-world settings (in the wild scenarios) like autonomous driving, there is a research gap for datasets that can help analyze and understand model limitations.

\subsection{Multi-modal Behavioral Datasets} 
Deep learning models may display unanticipated biases, prompting the creation of datasets designed to uncover their behavioral patterns \cite{abbas2023progress,alcorn2019strike,hosseini2018semantic,dong2020benchmarking,hendrycks2021natural}. Stylized ImageNet (SIN) \cite{geirhos2018imagenet} and IllusionBench \cite{hemmat2025hidden} investigate shape bias by introducing conflicts between texture and shape through synthetically generated illusory images. Similarly, several datasets have been developed to specifically study camouflaged or concealed objects \cite{fan2020camouflaged, le2021camouflaged, fan2021concealed, lv2021simultaneously, skurowski2018animal}. WHOOPS dataset \cite{bitton2023breaking} comprises synthetic images depicting unusual scenarios, requiring models to recognize anomalous elements. However, these images often necessitate additional contextual knowledge beyond generic object recognition (Fig.\ref{fig:whoops-example}). Multimodal Visual Patterns (MMVP) benchmark exploits differences in CLIP and DINOv2 embeddings to identify seemingly simple queries that lead to failure in multimodal large language models (MLLMs) \cite{tong2024eyes}. While several studies have investigated specific capabilities—such as fine-grained classification \cite{kim2024finer,chandhok2024response}, spatial arrangement \cite{chandhok2024response}, state of object \cite{gaur2024detect}, camera angle \cite{gaur2024detect}, color \cite{yiu2024kiva}, rotation \cite{yiu2024kiva,gaur2024detect}, reflection \cite{yiu2024kiva}, size transformations \cite{yiu2024kiva}, relative positioning \cite{gaur2024detect}, background \cite{gaur2024detect}, binding problem in feed-forward processing \cite{campbell2025understanding} and counting of objects \cite{chandhok2024response,yiu2024kiva,gaur2024detect}—these investigations primarily assess specific capabilities rather than their impact on improving visual perception or their role in logical processing in visual perception.

While existing studies and datasets assess various capabilities and reveal specific vulnerabilities, they do not enable a direct comparison of processing in human vision and deep learning models. Such a comparison is crucial for understanding the factors contributing to human perception's generalizability and robustness. As deep learning models increasingly achieve human-level performance on quantitative metrics, there is a need for a diverse and challenging dataset that can effectively benchmark these models and provide insights into their inductive biases.


%% file: sec/3_dual_thinking.tex
\section{Dual Thinking Framework}
\label{sec:dualthinking}

In this section, we discuss the computational advantages of using a dual-thinking framework for visual processing and the significance of existing theories on human vision within this framework.

\textit{\textbf{Theorem} : If we have multiple possible groupings involving a common region, with different prior probabilities, the lowest average number of iterations to reach correct grouping or segmentation is achieved by completely assigning the component to highest probabilities sequentially.}

Let $p_1, p_2, p_3..p_n$ be the probability in the order of most probable grouping, such that $p_1>p_2$ and $p_2>p_3$ and so on and there are $n$ possible groupings. The most probable grouping will be correct in the first iteration for $p_1$ fraction of the time, and the next highest probable grouping will take two iterations and be correct $p_2$ fraction of the time, and so on. The average length is given by 

\begin{equation} \label{eq:terms}
L_{avg} = 1p_1  +  2p_2 +  3p_3 ... +  np_n = \sum_{i=1}^{n} ip_i 
\end{equation}

We can prove the theorem by contradiction by showing that deviation from the condition in theorem increases the average length. Let us consider a case where a combination with probabilities $p_x$  and $p_y$ (such that $p_x > p_y$) are assigned position $j$ and $i$, where $j > i$ to prove the theorem by contradiction. In this case the higher probability combination is tried out a at later stage. The difference in average length in this case compared to correct assignment would be $(jp_x + ip_y) - (ip_x + jp_y)$ = $p_x(j-i) - p_y(j-i)$ = $(p_x-p_y) (j-i)$. Since $p_x > p_y$ and $j > i$, this will results in a positive quantity. As deviating from the condition stated in the theorem causes an increase in the average length, the best possibility is the condition stated in the theorem. 

We can also note that the $\sum_{i=1}^{n} p_i = 1$ and when $p_1 \to 1$ other  $p_i$  tends to 0 for $i \neq 1$.

\begin{equation} \label{eq:avglen}
\lim_{p_1 \to 1} L_{avg} \to 1 
\end{equation}

The intuitive processing in human vision is quite accurate; hence, we usually do not perceive the first inference. However, multiple inferences are perceivable in images in our dataset as the first inference is wrong and differs from the final inference. For example, in Fig.\ref{fig:human-fg} referenced later, interpreting the shadow as a skirt worn by the second lady is the human intuitive perception, and ignoring the shadow and considering only the second lady is the correct perception.

\paragraph{Sub-component Formation}
Let us consider a group of pixels $\mathbf{I}$ = \{${i_{x,y}}$\} where ${i_{x,y}}$ represent pixel at ($x$, $y$) position. Instance segmentation creates disjoint and nonempty subsets of pixels ($s_m$) that correspond to object instances or the background and satisfy the conditions that each pixel can belong to only one instance, and the segments must collectively cover the entire image (eq. \ref{eq:non-exclusive}).

\begin{equation} \label{eq:non-exclusive}
S = \{ s_0, s_1, .. s_T\}  
\quad 
where \quad s_m \cap s_n = \emptyset \quad and \quad \bigcup\limits_{m=0}^{T} s_m = I \nonumber\\
\end{equation}

\begin{equation} \label{eq:best-guess}
c_j = \{ i_{x,y}\}  \quad where \quad \bigcup\limits_{j} c_j \quad \cup s_0= I \quad and \quad c_j \cap c_k = \emptyset
\end{equation}

In segmentation, the complexity or hardness is informally considered as the size of the search space \cite{adolfi2022computational,adolfi2022computationalcogsci,franklin2020structured}, and the number of possible instances or the subsets possible for the set $\mathbf{I}$ is $O(2^{K})$, where 
$K $ is the number of pixels in the image. A common approach is to reduce the number of elements \cite{brent1999speech,friston2021active}. We can observe that figure-ground extraction can reduce a significant number of pixels by ignoring the background ($s_0$). The formation of sub-components ($C$) using Gestalt principles (e.g., proximity and similarity) or edges can reduce possible combinations from $O(2^{K})$ to $O(2^{card(C)})$. This sub-component formation is also amenable for iterative processing as rectification of incorrect grouping in pixel-based processing can result in redundant intermediate steps observed in segmentation models that use iterative processing \cite{wang2019recurrent,tomar2022fanet}, and prevent partial sub-component assignments (Fig.\ref{fig:human-logical-amodal-gaps} in the results section) or different types of boundary errors \cite{cheng2021boundary}.

\paragraph{Instance and Class Identification} 
The shape can help evaluate the possibility of the groupings in this search space to form an instance (e.g. in Fig.\ref{fig:human-logical-impossible}, with more probable shapes that correspond to common postures can be prioritized for faster computation in intuitive processing (similar to Theorem 1), and can be evaluated till $\bigcup\limits_{j} c_j$ $\cup $ $s_0= I$ is satisfied (e.g. Fig.\ref{fig:human-logical-impossible}). The space of all possible groupings of components can be searched optimally to identify instances using shape as it prioritizes sub-groups that fit the different object shapes or postures. 

\paragraph{Logical Processing}
Evaluating all possible combinations is computationally infeasible (eq.\ref{eq:terms}). Instead, combinations can be examined iteratively until no errors are found in the generated instances. The absence of errors in identified instances can serve as an early stopping criterion, improving energy efficiency in human vision. Processing time can be further optimized by using insights from previous evaluations to reorder subsequent assessments based on updated probabilities or by limiting the search to predefined sub-groups or sub-regions (eq.\ref{eq:avglenimprove}).

\begin{equation} \label{eq:avglenimprove}
L_{avg} = 1p_1  +  2p_2' +  3p_3' ... +  np_n' = p_1 + \sum_{i=2}^{n} ip_i' 
\end{equation}

Early stopping criteria can enhance the energy efficiency of human vision; however, they may also allow specific errors, undetected by logical processing, to persist. Recent research on System 1 and deliberate System 2 analysis emphasizes that users consciously examine each region \cite{buccinca2021trust} to achieve complete image coverage in critical domains like medical imaging \cite{gefter2022special}. Our study examines errors corrected through logical processing and those that persist in the results section.

%% file: sec/4_methods.tex
\section{Methods}
\label{sec:methods}

In the first part, we introduce Human confusion dataset consisting of images adversarial for human vision to study dual thinking framework and deep learning models. We then propose a method to automatically evaluate model correctness and similarity to intuitive processing in human vision. 

\subsection{Human confusion dataset} 

\begin{figure}[ht]
\centering
    \subfloat[ ]{\includegraphics[height = 1.8cm]{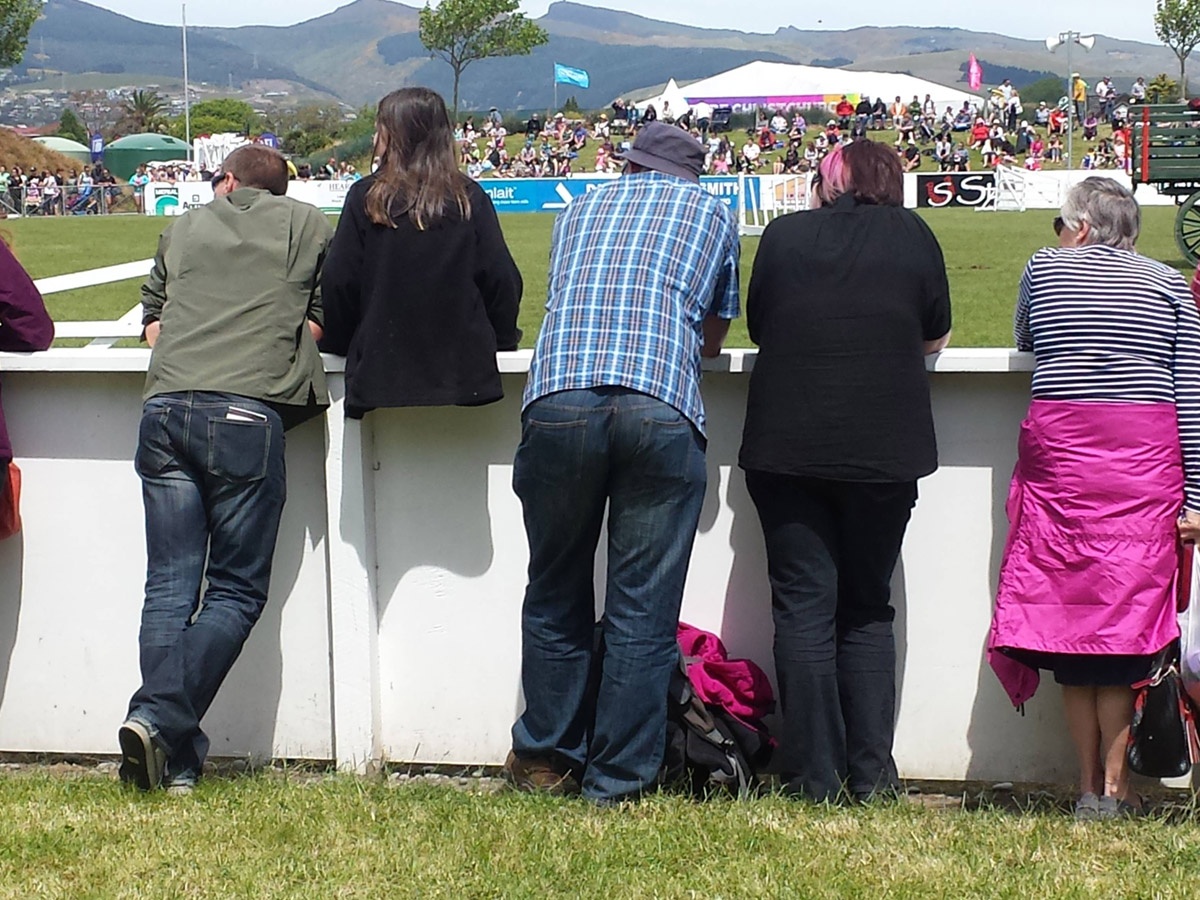}  \label{fig:human-fg} }      
    \subfloat[ ]{\includegraphics[height = 1.8cm]{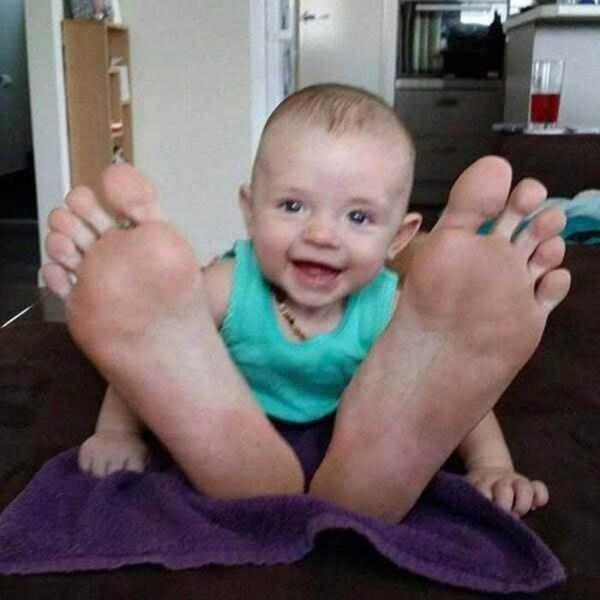} \label{fig:human-proximity} }  
    \subfloat[ ]{\includegraphics[height = 1.8cm]{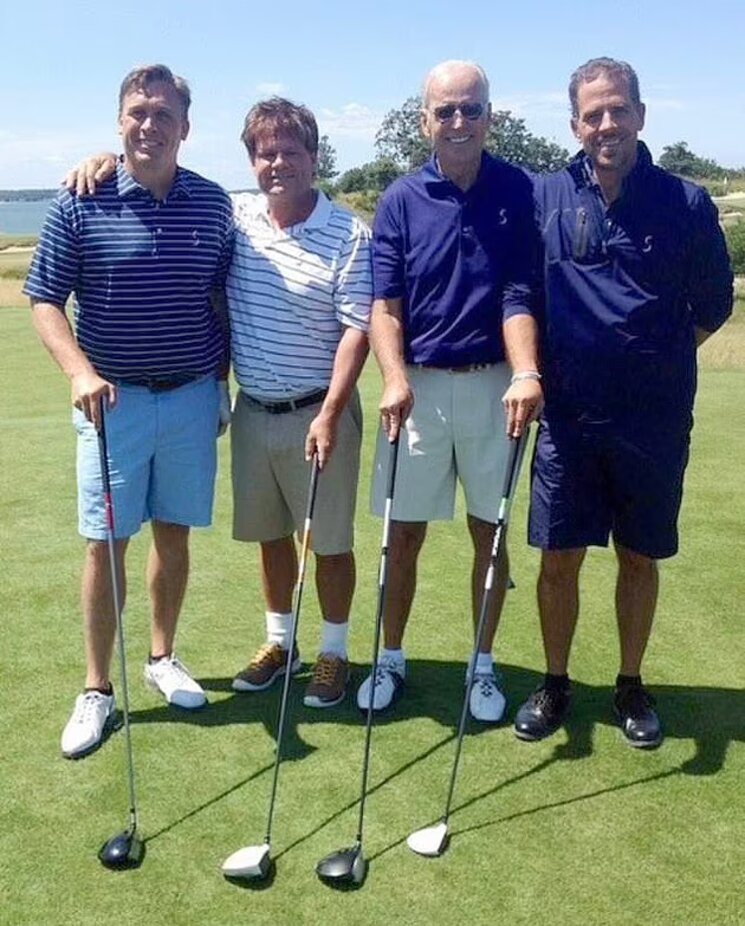} \label{fig:human-similar} }        
    \subfloat[ ]{\includegraphics[height = 1.8cm]{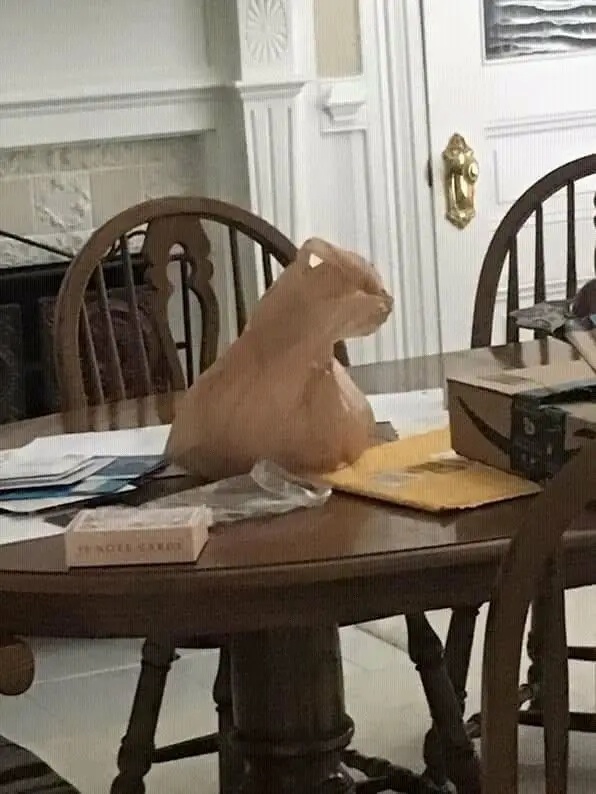} \label{fig:human-fg2} }    
    \subfloat[ ]{\includegraphics[height = 1.8cm]{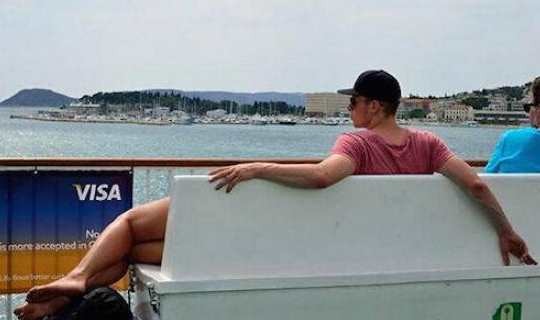} \label{fig:human-amodal}}    
    \subfloat[ ]{\includegraphics[height = 1.8cm]{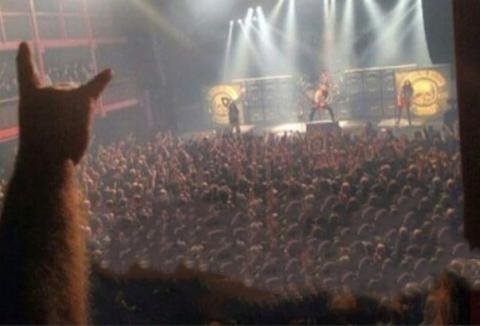} \label{fig:human-logical-label1}}        
\caption{(a) Figure ground with a part of background grouped with the second lady (b) Proximity and similarity error where the legs are grouped with the child. (c) Similarity, continuity and proximity error occurring together, the shade on hand makes it similar to adjacent shirt and forms a continuous smooth curve. (d) Figure ground error where cover is wrongly identified as cat (e) Amodal Closure and Similarity error where components from two instances that are apart grouped together as a single instance. (f) An example of image that are not analyzed using segmentation models as the confusion is about class of identified instance}    
\label{fig:modelbar}
\end{figure}

Grouping elements with similar properties, including brightness, color, contrast, texture, etc, is called similarity (e.g., Fig.\ref{fig:human-similar}). Continuity is important in human vision and is concerned with grouping elements that form smooth contours (e.g Fig.\ref{fig:human-similar}), and proximity refers to closer elements likely to be grouped \cite{wagemans2012century} (e.g Fig.\ref{fig:human-proximity}). Human vision focuses only on salient regions of the image (or foreground) for analysis, and an example of a figure-ground error is shown in Fig.\ref{fig:human-fg} and \ref{fig:human-fg2}. Amodal closure denotes the completion of occluded elements \cite{bnanay2018amodal,murray2004setting}, and errors mainly occur when we perceive parts of different instances as a single instance (e.g., Fig.\ref{fig:human-amodal}). The errors can also be due to a combination of many properties, for example, similarity, proximity, and continuity in Fig.\ref{fig:human-similar} and similarity and proximity in Fig.\ref{fig:human-proximity}. A comprehensive review of Gestalt principles is provided in \cite{wagemans2012century,wagemans2012century2}; and examples in supplementary data. 

\subsubsection{Data Collection} 
We crowd-source the initial data using Amazon Turk, where participants collect 150 confusing images by searching for "confusing photos", "images that need to be looked twice to understand," and similar terms. The initial three participants collected generic images, and the following four participants collected confusing images related to common objects. To expand the dataset, we engaged three additional participants to collect confusing images that were not part of the crowd-sourced set, as there was significant overlap and many synthetic images among the collected data using Amazon Turk. In total, ten participants contributed to the image collection, including seven from Amazon Turk and three for dataset expansion. 

\subsubsection{Data Preparation} 
We remove duplicate and synthetic images from the collected dataset and filter images where the instance segmentation task cannot identify the confusion or the confusion pertains to objects that do not belong to classes in the MS-COCO dataset. The final dataset consists of 1000 images without any NSFW content in correct perception with confusing instances belonging to classes in MS-COCO dataset \cite{lin2014microsoft}. Using open-clip embeddings, we identified over 350 images matching entries in the LAION-5B dataset \cite{schuhmann2022laion} (NeurIPS-2022), making them available for research. Annotations include details about image presence in existing datasets and those available in the open domain. We have shared the URL's of all 1000 images in the dataset. 

We annotate each image on \textbf{(a)} Gestalt principles (figure-ground, proximity, continuity, similarity, and amodal closure) except modal closure to analyze qualitative behavior, \textbf{(b)} logical errors (size and count difference) to evaluate vulnerabilities, \textbf{(c)} instances with global shape of one instance but containing instances of different class to evaluate shape bias, and \textbf{(d)} images in wild that can help understand out of distribution behavior. The annotation also includes the operations and masks to check the correctness and similarity to the intuitive perception. 

\subsection{Psychophysical Experiments}

We conducted experiments with 100 participants, who were shown the images in batches of 100 or 200 images, and information was sought on the confusion in the images. The first two questions—whether participants \textit{"found anything confusing in the image"}—helps us confirm that the images contained ambiguous regions, while the follow-up question—whether they \textit{"could immediately recognize the correct interpretation"}—allowed us to assess for multiple inferences and dual thinking. The questions in our study included: (a) Did you find anything confusing? (b) Did you immediately recognize it? (c) Would you have identified the confusion if the image had not been labeled as confusing? (d) For participants who initially found no confusion, we showed them the confusion masks and then asked if they found that confusing. These questions also allowed us to investigate early stopping in human visual processing. 

\subsection{Model Evaluation}

Multi-modal LLM's currently do not support instance segmentation and respond only in text. To evaluate LLM's, we frame questions that allow us to assess whether the model's response aligns with intuitive or correct perception. The questions had answers in Yes/No format or numeric format and prompts were used to instruct the LLM's to respond accordingly (\textit{"Answer only Yes or No"} or \textit{"Answer only in Numeric"}). The model outputs were then compared against annotated correct answers and intuitive perceptions.
Similarly, we evaluate the confusing region separately in segmentation models to prevent it from being overwhelmed by the overall error \cite{cheng2021boundary}. In our analysis of images in the dataset, we observe that the errors in human perception mainly fall into the following categories \textbf{(a)} an object is missed out completely, \textbf{(b)} a part of an instance or background is wrongly grouped with another instance \textbf{(c)} two instances group to form a single instance \textbf{(d)} a part of the background identifies as an object or \textbf{(e)} an instance is misclassified to be of a different class. 
We identify the confusing instance using the Intersection over Union ($IoU$) score with the instance mask and the model outputs to identify confusing objects. We use separate masks to identify human perception and correct perception to make evaluation independent of the size of the confusing region. The algorithm to evaluate the different errors is shown in Alg.\ref{alg:CIF} and more details are present in the supplementary data. 

\begin{algorithm} 
 \caption{Basic automatic evaluation algorithm}
 \begin{algorithmic}[1]
 \renewcommand{\algorithmicrequire}{\textbf{Input:}}
 \renewcommand{\algorithmicensure}{\textbf{Output:}}
 \REQUIRE modelOutput, instanceMask, confusionMask, includeConfusion, cThreshold
 \ENSURE  correctStatus
  \STATE matches $\gets$ [] 
  \FOR {each $instance$ $\in$ modelOutput}
  \STATE score $\gets$  IoU($instance$, instanceMask)   
  \IF {(score \textgreater match\_threshold)}
  \STATE  matches.insert(($instance$,score))
  \ENDIF
  \ENDFOR
  \STATE matches.sort(key=score,reverse=True)
  \STATE result = $False$
  \FOR {each $match,score$ $\in$ matches}
  \STATE overlap $\gets $ IoU(($match$*confusionMask,\hspace{0.25em}confusionMask)
  \IF {($overlap$ \textgreater\hspace{0.25em}cThreshold) $\AND$ includeConfusion}
  \STATE  result = $True$
  \ENDIF
  \IF {($overlap$ \textless\hspace{0.25em}cThreshold) $\AND$ $\NOT$ includeConfusion}
  \STATE  result = $True$
  \ENDIF
  \ENDFOR
 \RETURN {} result
 \end{algorithmic} 
 \label{alg:CIF}
 \end{algorithm}

\subsection{Experimental Setting}
In our study, we consider eleven multi-modal LLM's, GPT-4o mini \cite{openaigpt4omini}, GPT-4o \cite{hurst2024gpt} and reasoning model O3 from OpenAI and open source model LlAMA 3.2 11B and LlAMA 3.2 90B \cite{metallama} from Meta, Deepseek-VL from Deepseek, Qwen2.5-VL-7B-Instruct and Qwen2.5-VL-72B-Instruct \cite{bai2025qwen2} from Alibaba, Gemini-2.0-Flash \cite{google2024gemini2} from Google, Sonnet-4 from Anthropic \cite{anthropic2025claude4} and Pixtral-12B \cite{agrawal2024pixtral} from Mistral AI. We use the API from the respective companies for models from OpenAI, Anthropic and Google, we use API from Hyperbolic for Qwen2.5-VL-7B-Instruct, Qwen2.5-VL-72B-Instruct and Pixtral-12B. We had setup LlAMA and Deepseek-VL (version 2) locally for our experiments. While O3, GPT-4o mini, GPT-4o, Sonnet-4 and Gemini-2.0-Flash were closed, LlAMA, Qwen2.5-VL and Pixtral-12B were open weight models. 

We use segmentation models from MMDetection framework \cite{chen2019mmdetection}, trained on the MS-COCO dataset \cite{lin2014microsoft}, and consider model outputs above the confidence threshold of $0.7$.  We evaluate models on $983$ images except for $17$ images with confusion related to the labels (e.g., Fig.\ref{fig:human-logical-label1}). We select the thresholds for the automatic evaluation by comparing human annotations on a randomly selected subset of $250$ images from our dataset on seven models. The threshold for recognizing instance is $73\%$, $25\%$ overlap of confusion region not part of an instance, and $70\%$ for being part of an instance. The automatic detection algorithm matched the human annotation decisions in $94.94\%$ of cases. We randomly selected 1,000 images from the MS-COCO dataset and generated questions similar in style to those in our Human Confusion Dataset. We then evaluated three models—4o-mini, Qwen-7B, and sonnet-4—on these questions. The models achieved higher accuracy on the MS-COCO data (93.8\%, 92.9\%, and 88.9\%, respectively) compared to their overall accuracy on our dataset (70\%, 65\%, and 62\%).    

%% file: sec/5_results.tex
\section{Results}
\label{sec:results}

In this section, we provide evidence for the dual thinking framework in human vision and discuss different strategies used in these stages. We then analyze deep learning models using the method described earlier. 

\subsection{Human Visual Processing}

In this section, we first examine the evidence for dual thinking and aspects of intuitive and logical processing using our Human Confusion dataset. We then present the results of psychophysical experiments and demonstrate the usefulness of our dataset to study dual thinking in human vision.

\begin{figure}[ht]
\centering      
    \subfloat[Optical illusions are usually shortcomings of human vision (Ebbinghaus, Müller-Lyer, Café Wall, Checker Shadow Illusion. \label{fig:human-illusion}] {
        \frame{\includegraphics[width = 1.8cm]{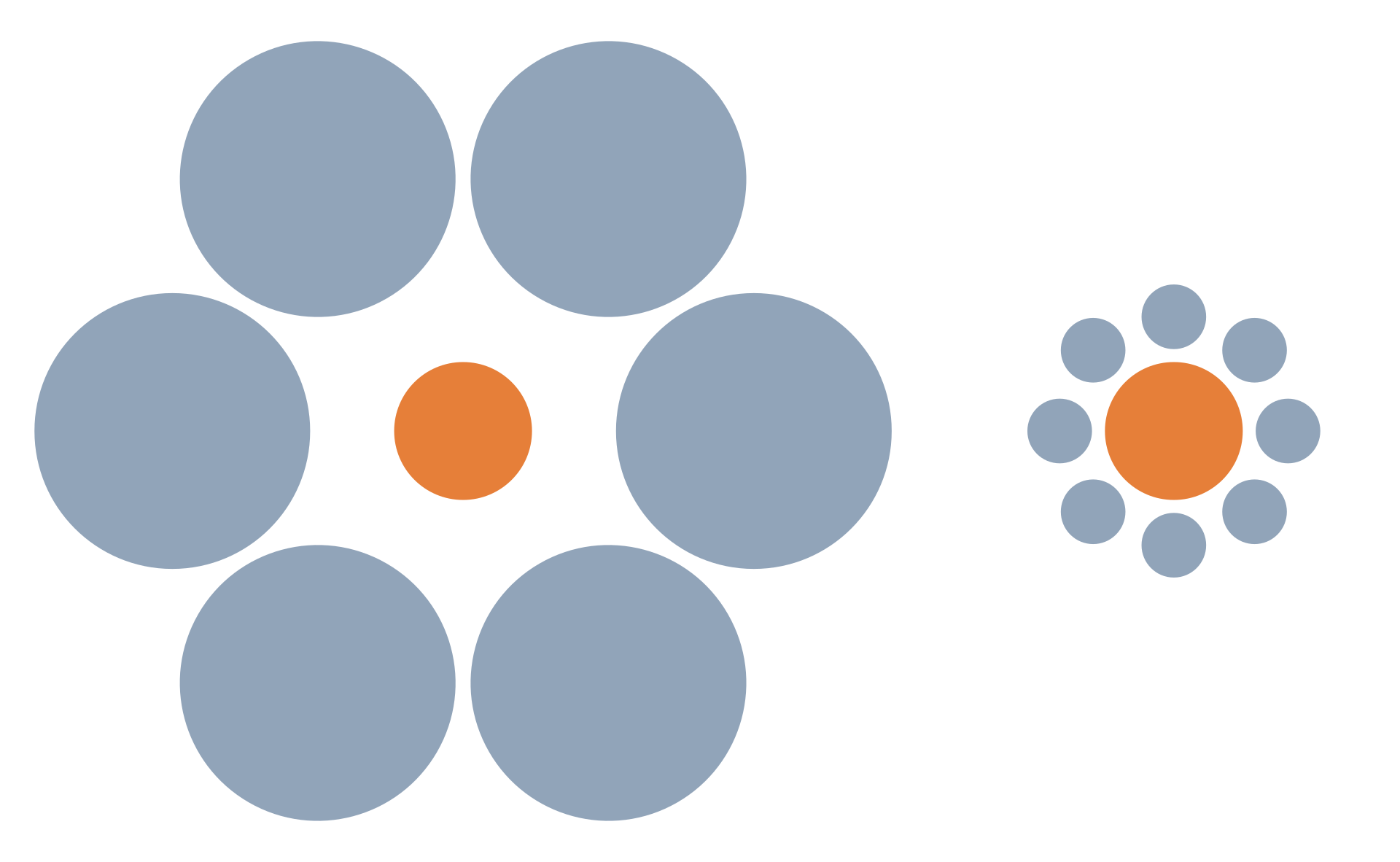}}
        \frame{\includegraphics[width = 1.85cm]{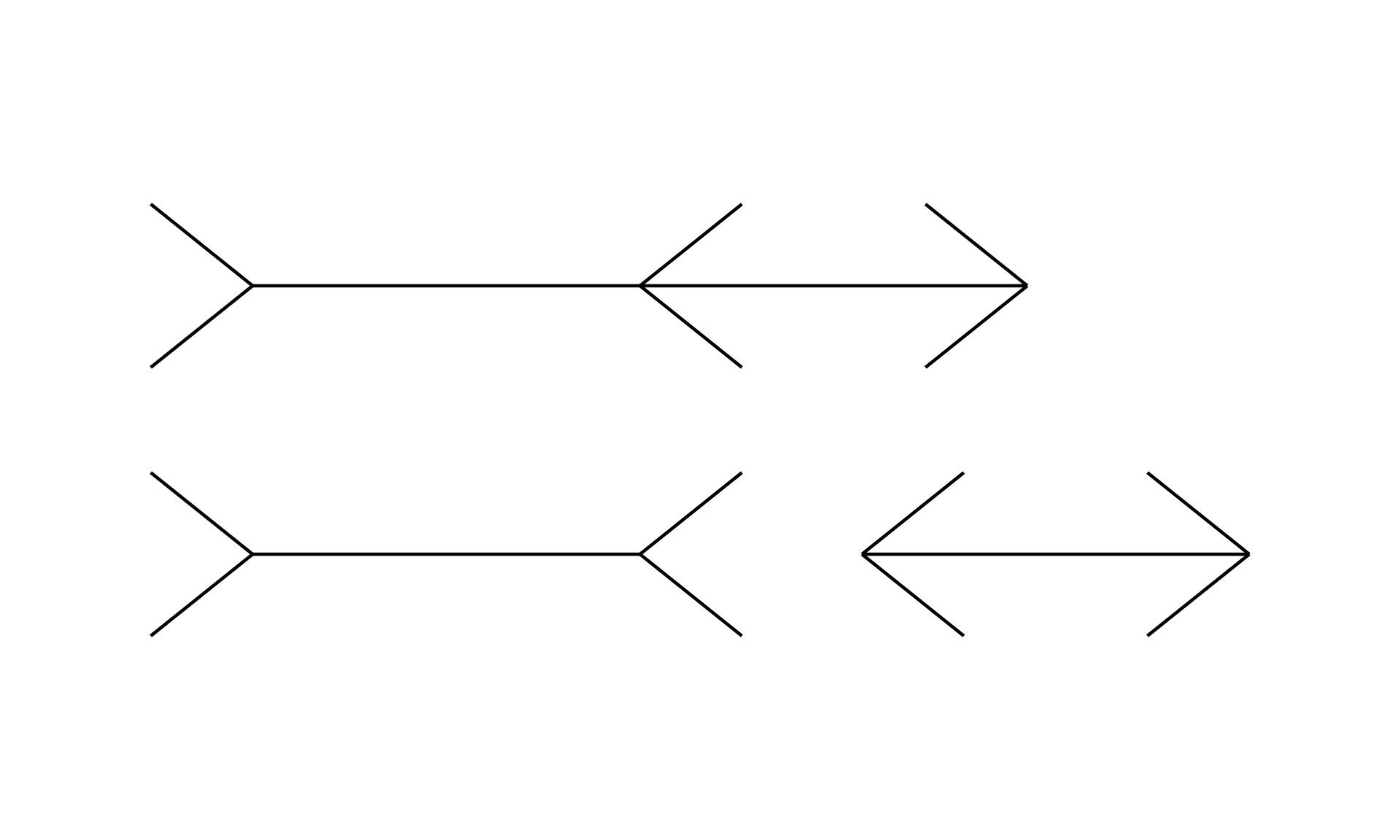}}
        \frame{\includegraphics[width = 1.725cm]{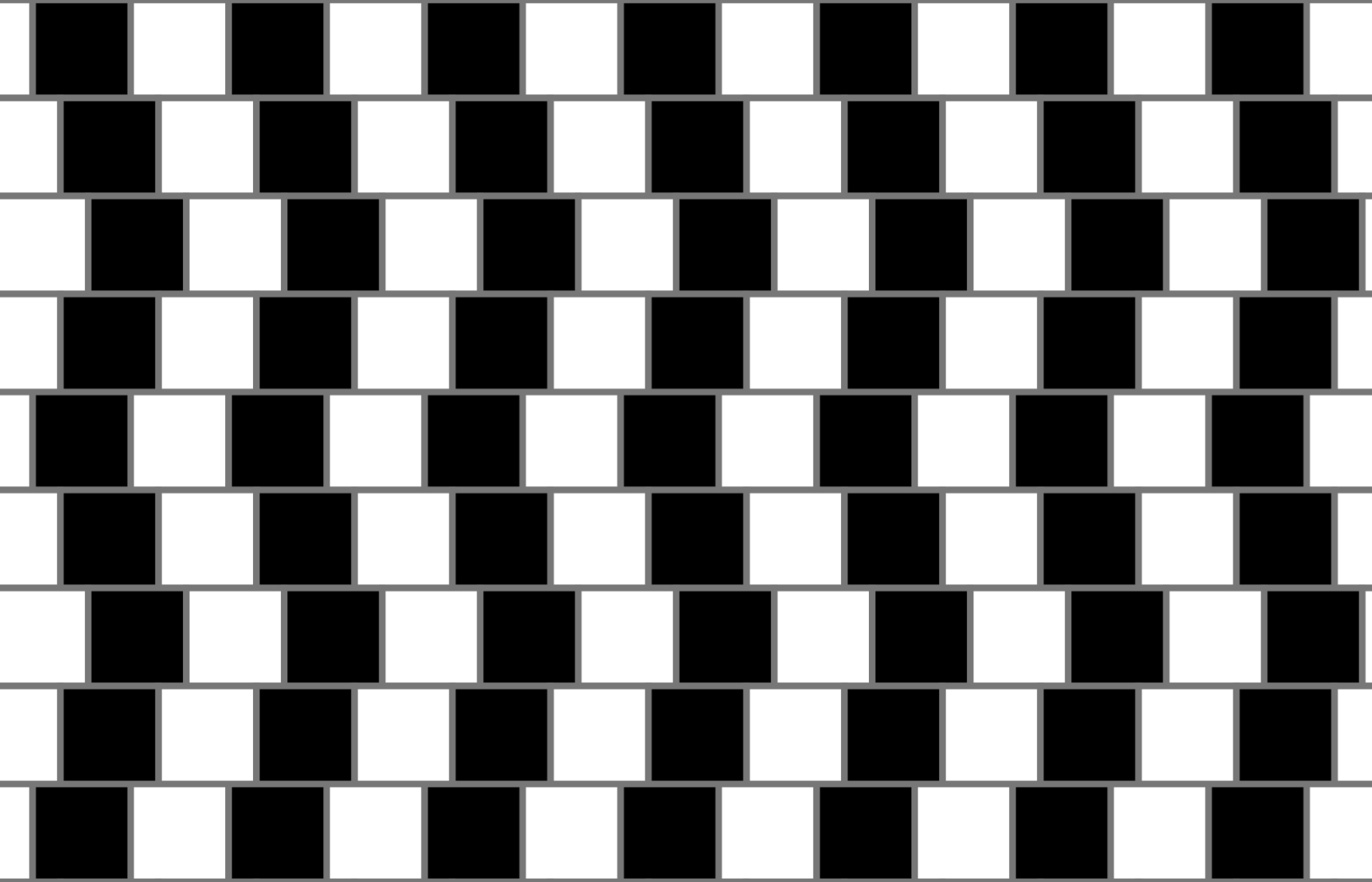}}
        \frame{\includegraphics[width = 1.45cm]{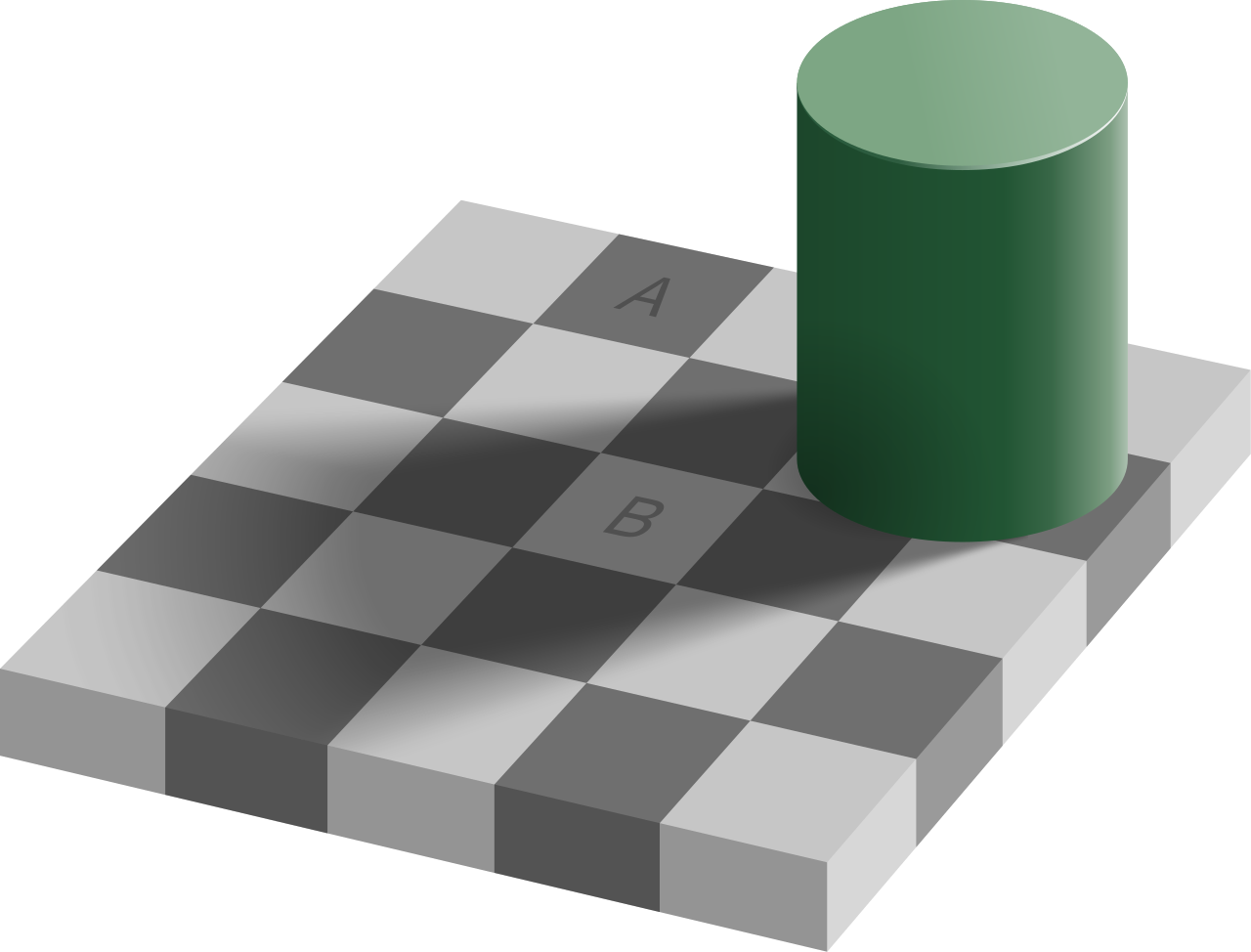}}
    } \hfill 
    \subfloat[In dual thinking the false perception in intuitive processing is easily identified and immediately corrected. \label{fig:human-dog-mix}] {
        \includegraphics[height = 1.4cm]{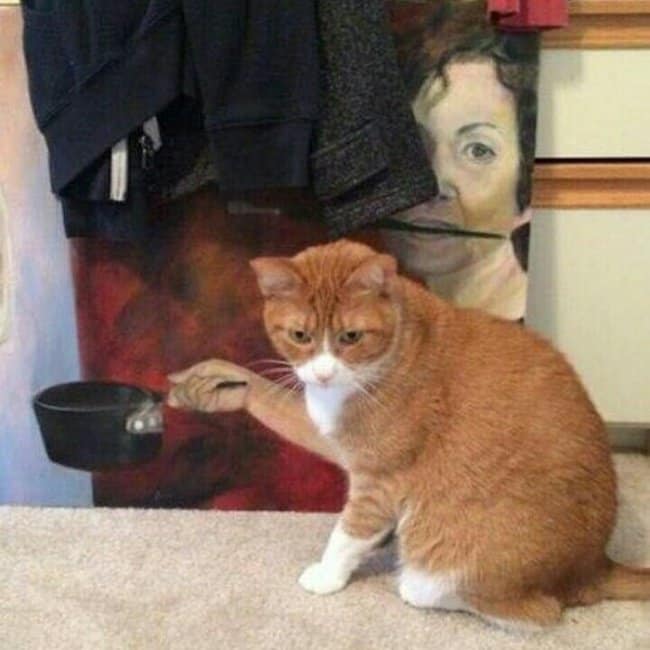}
        \includegraphics[height = 1.4cm]{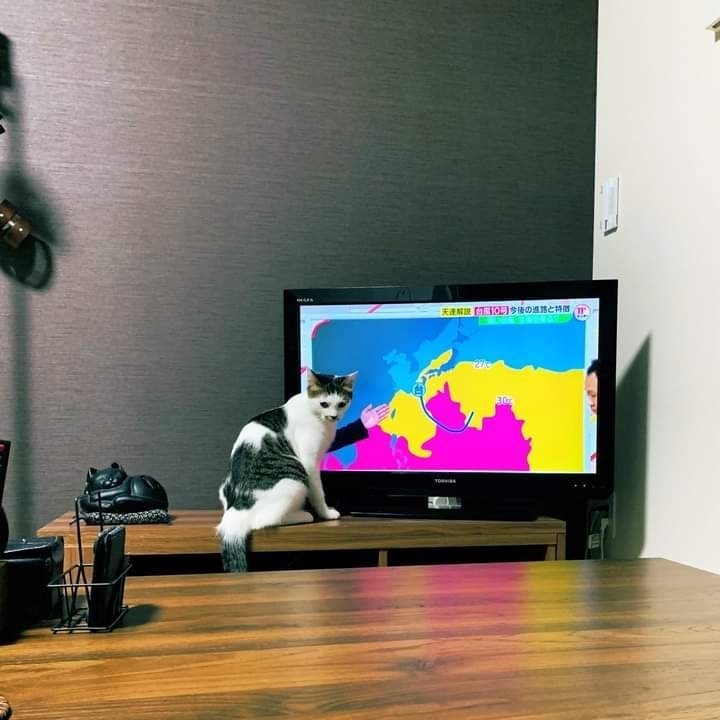}
        \includegraphics[height = 1.4cm]{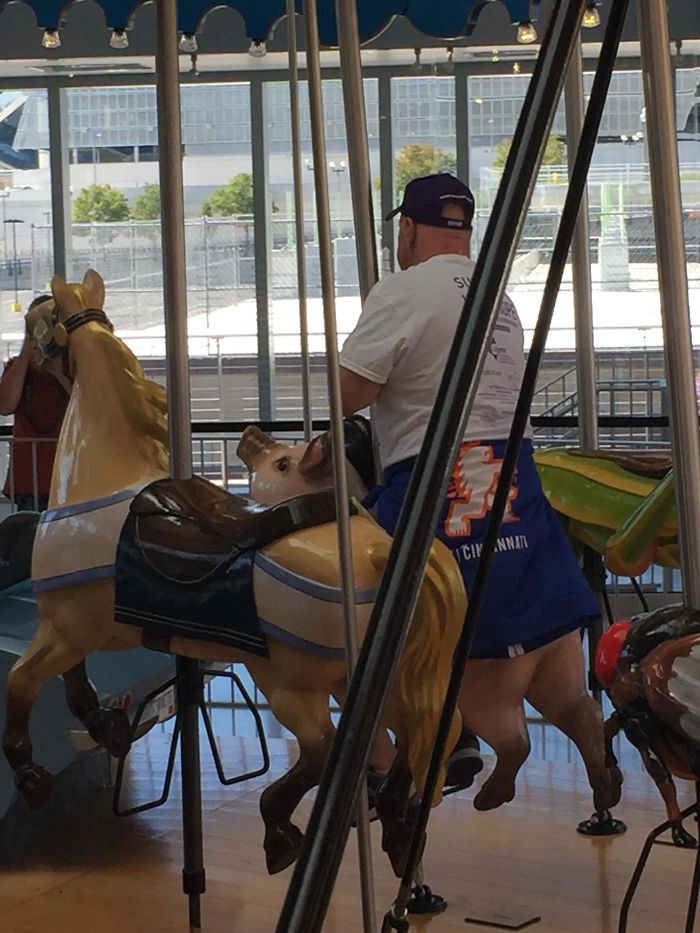}
        \includegraphics[height = 1.4cm]{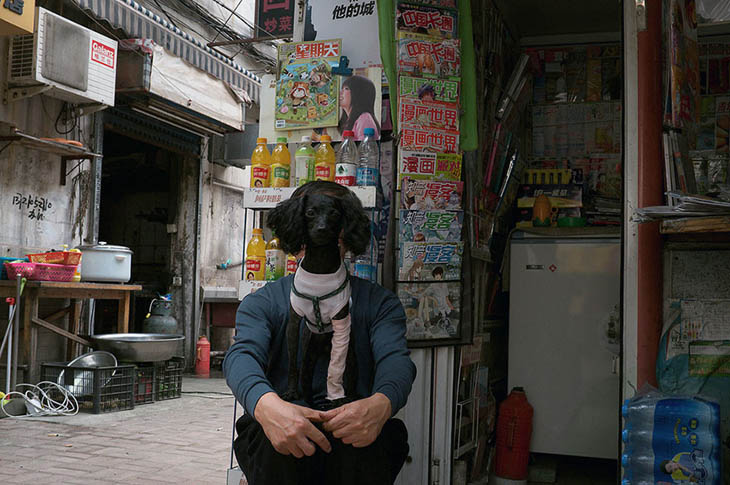}           
    } 
    
    \subfloat[Shape can group components even without specific regions like eyes in intuitive perception. \label{fig:human-mix-shape}] {
        \includegraphics[height = 1.42cm]{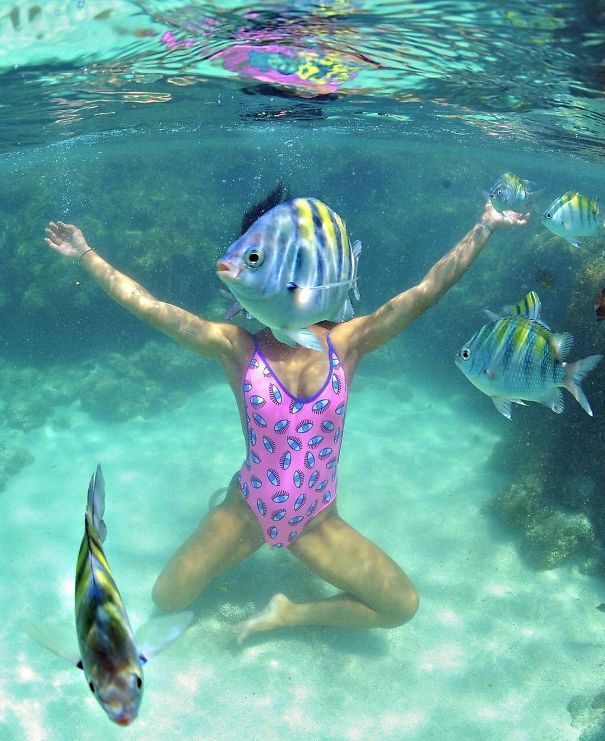}
        \includegraphics[height = 1.42cm]{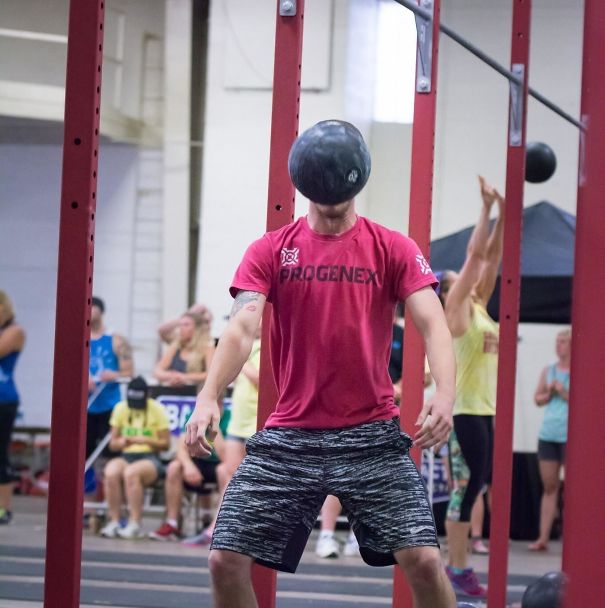}
        \includegraphics[height = 1.42cm]{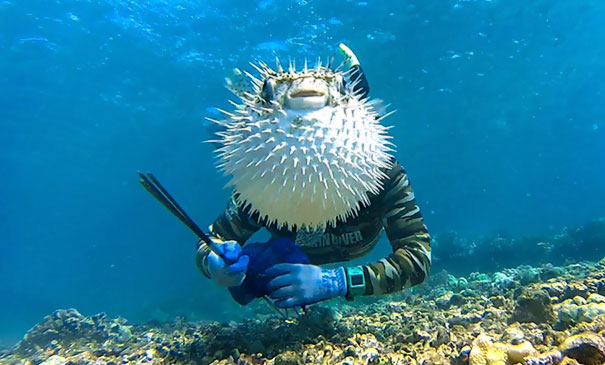}
        \includegraphics[height = 1.42cm]{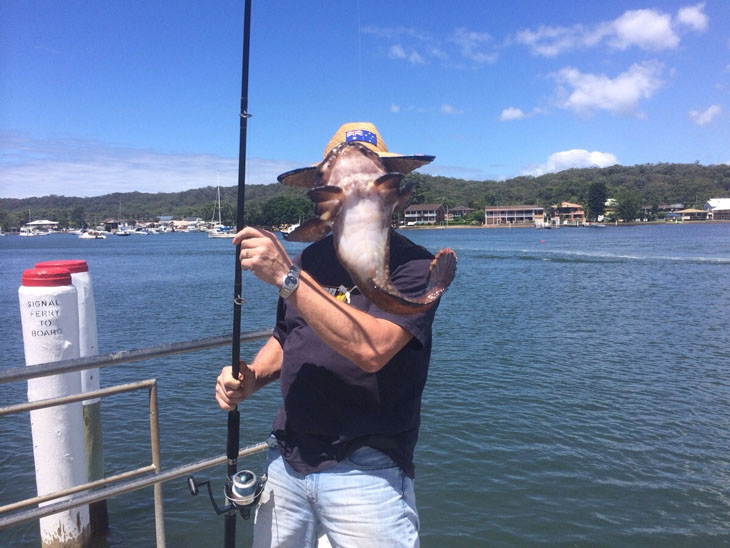}            
    } \hfill 
    \subfloat[Shape is an important factor in grouping sub-components. \label{fig:human-dog-mix}] {
        \includegraphics[height = 1.45cm]{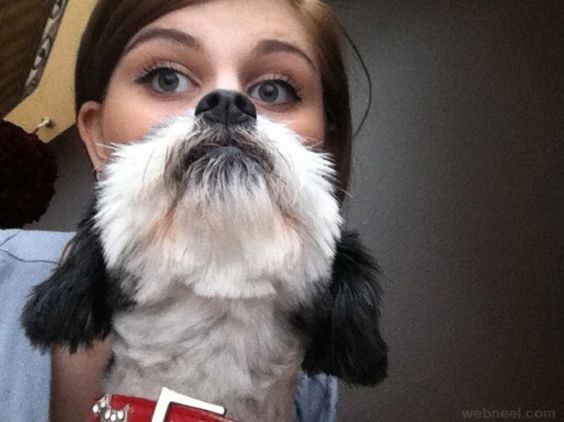}
        \includegraphics[height = 1.45cm]{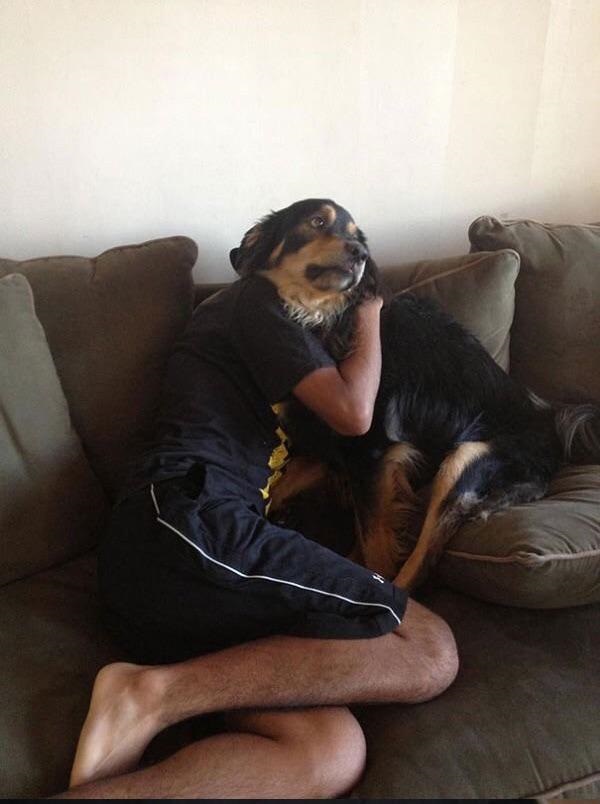}
        \includegraphics[height = 1.45cm]{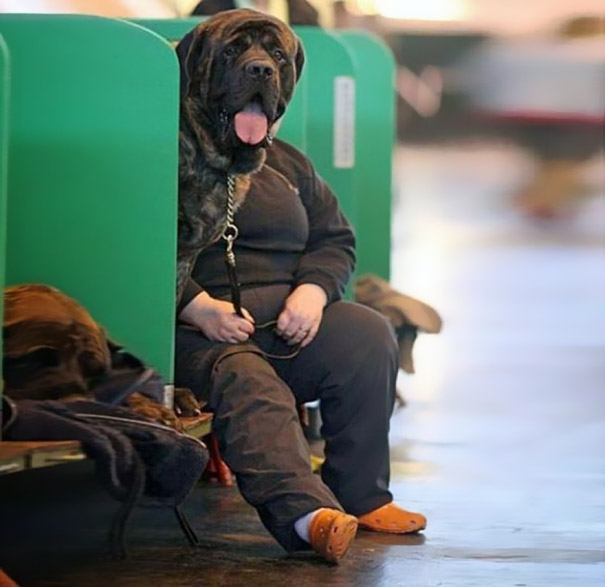}
        \includegraphics[height = 1.45cm]{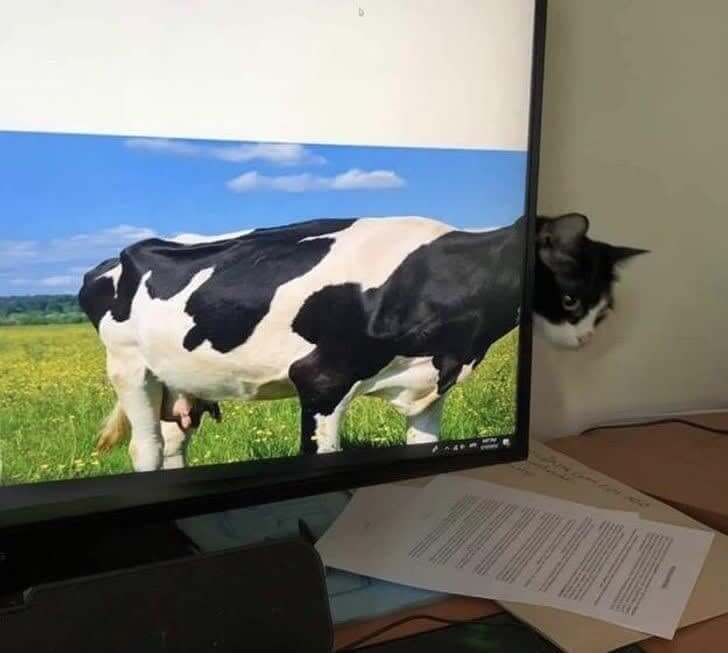}            
    }
\caption{Multiple inferences indicates dual thinking framework.}
\label{fig:shape-segm}
\end{figure}

\paragraph{Dual thinking framework}
 We could perceive multiple inferences, including the wrong first inference followed by a correct inference in our dataset, indicating the presence of a dual-thinking framework. The initial inference is corrected almost immediately (e.g., Fig.\ref{fig:human-dog-mix},\ref{fig:human-mix-shape}), which is in contrast to optical illusions used to demonstrate the shortcomings of human vision, where the decision does not usually change (Fig.\ref{fig:human-illusion}). In some images, the logical processing does not identify the errors; however, the users can recognize them retrospectively, which validates a common assumption in computational models of human vision about processing till a confidence threshold is attained \cite{tscshantz2023hybrid}. For example, the camouflaged bird in Fig.\ref{fig:human-logical-camo-bird} is not easily identifiable, but as these are not limitations of human vision, the users can identify the bird when mentioning the location. 

\noindent
\textbf{Instance and Class Identification} 
Shape as a method for grouping components to form instances is observable in grouping errors due to common posture (e.g. Fig.\ref{fig:human-logical-impossible}), and the examples where two components from two instances with the shape of an instance (e.g., Fig.\ref{fig:human-mix-shape}). The priority for shape over color for grouping components in intuitive processing is observable in amodal closure errors, where sub-components of a single class with widely different colors are grouped as they have the shape of a single instance. The shape also had priority over length in grouping components that are widely separated. 
Our dataset provides evidence for the importance of shape in instance identification compared to existing studies that focus on shape for classification. We can observe the formation of sub-components, which is evident in errors that occur in groups of related pixels. Gestalt principles and edge-based processing can help in sub-component formation; however, the adversarial examples only indicate sub-component formation but cannot distinguish between these approaches.

\paragraph{Psychophysical Trials} Our analysis of psychophysical trials showed that participants found on average, 96\% of the images confusing. In 51.69\% images the confusion was immediately recognizable, indicating the errors in intuitive processing that are corrected immediately and can be considered examples of dual thinking. Participants also noted that, for many images, they might not have noted the confusing regions without prior mention as confusing images before the experiment. This suggests early stopping in human vision, where processing stops after the perception deems the information sufficient. 

\subsection{Segmentation Models}

\begin{figure}[ht]
\centering    
    \includegraphics[height = 1.25cm]{images/dataset/image_0325.jpg}  \label{fig:human-logical-count-0} \hfill
    \includegraphics[height = 1.25cm]{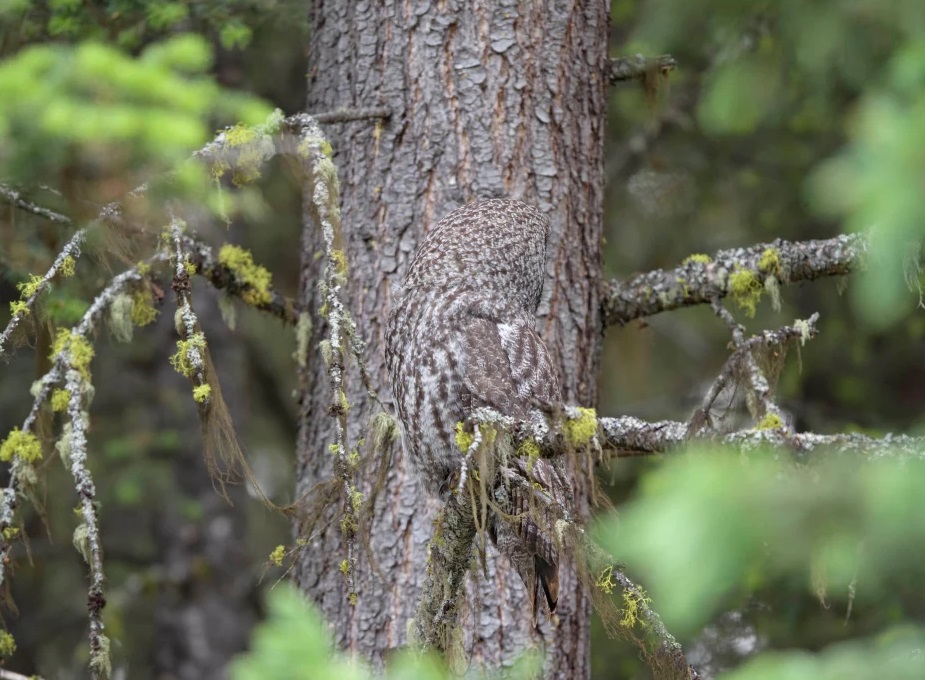} \label{fig:human-logical-size-0} \hfill
    \includegraphics[height = 1.25cm]{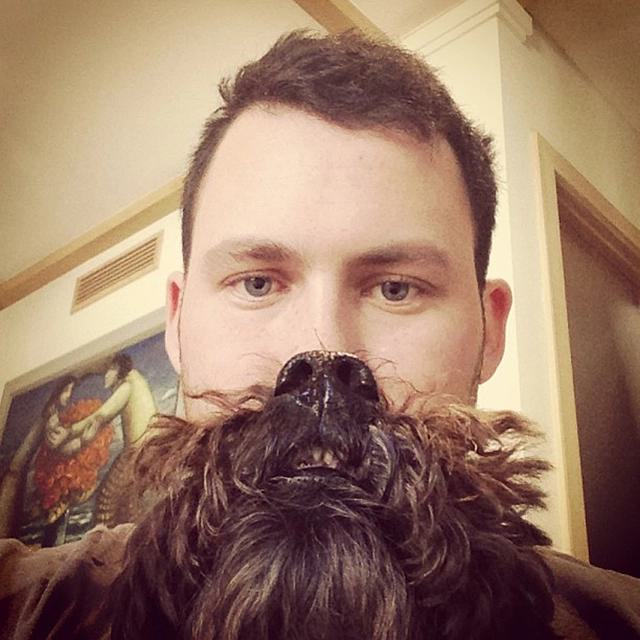} \label{fig:human-logical-impossible-0} \hfill
    \includegraphics[height = 1.25cm]{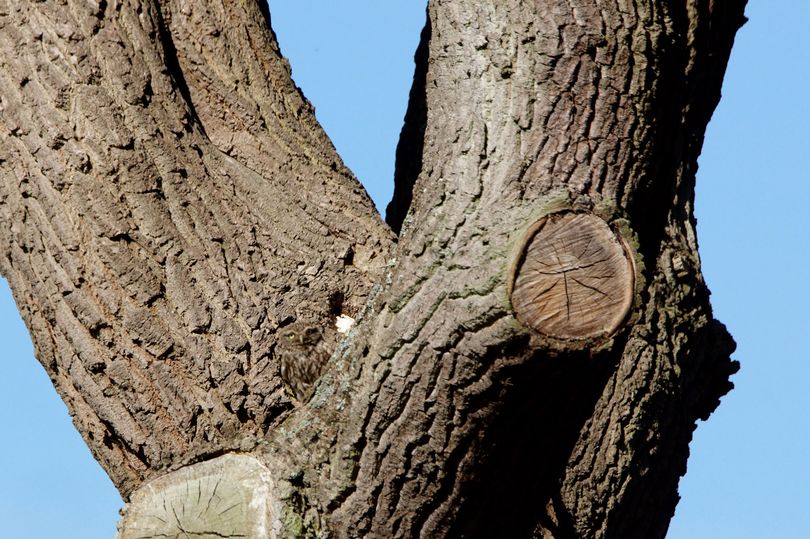} \label{fig:human-logical-amodal-0} \hfill
    \includegraphics[height = 1.25cm]{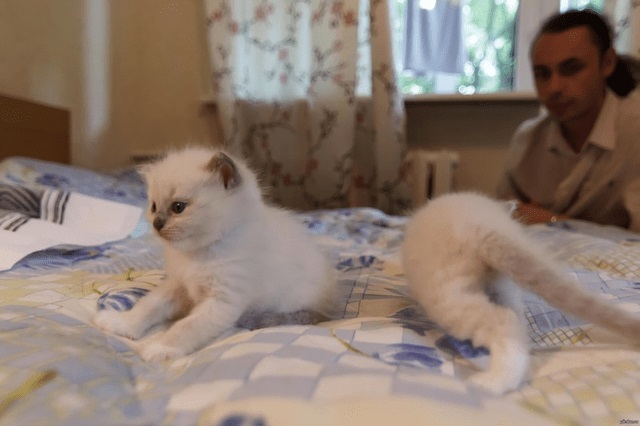} \label{fig:human-logical-amodal-0} \hfill   

    \includegraphics[height = 1.25cm]{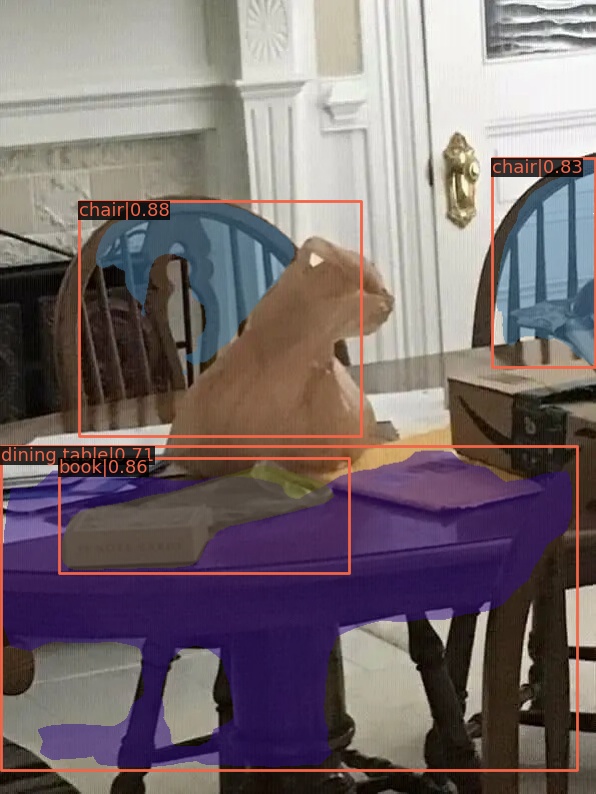}  \label{fig:human-logical-count-0} \hfill
    \includegraphics[height = 1.25cm]{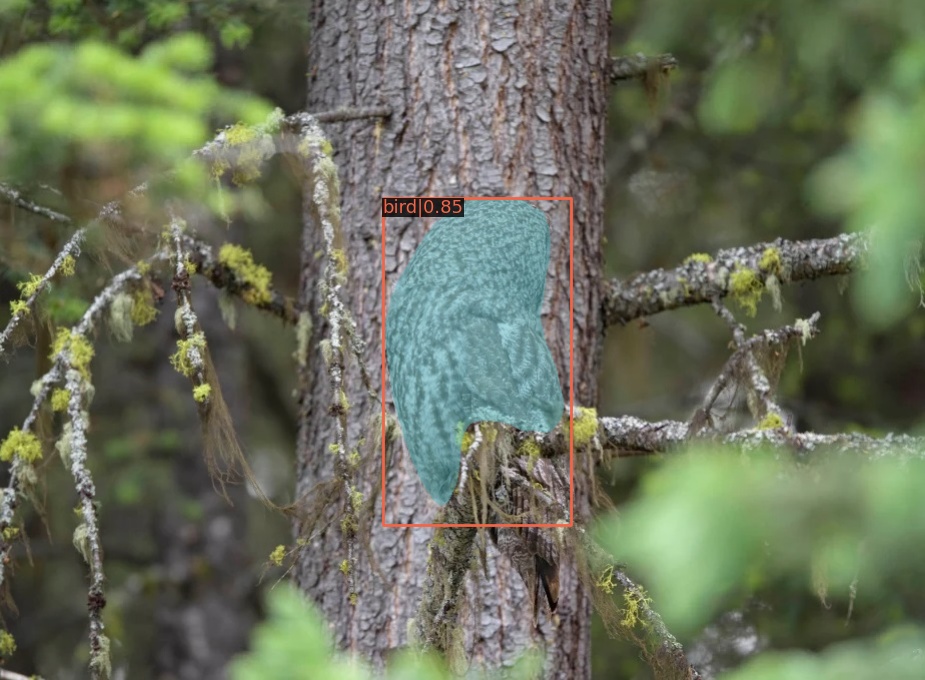} \label{fig:human-logical-size-0} \hfill
    \includegraphics[height = 1.25cm]{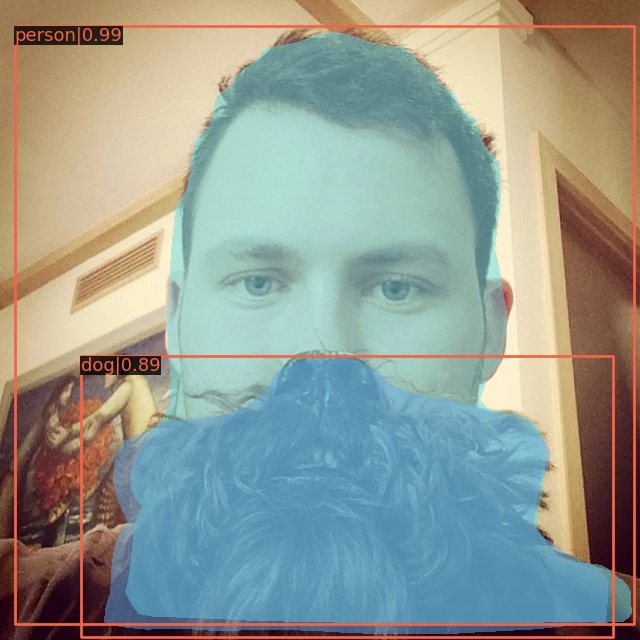} \label{fig:human-logical-impossible-0} \hfill
    \includegraphics[height = 1.25cm]{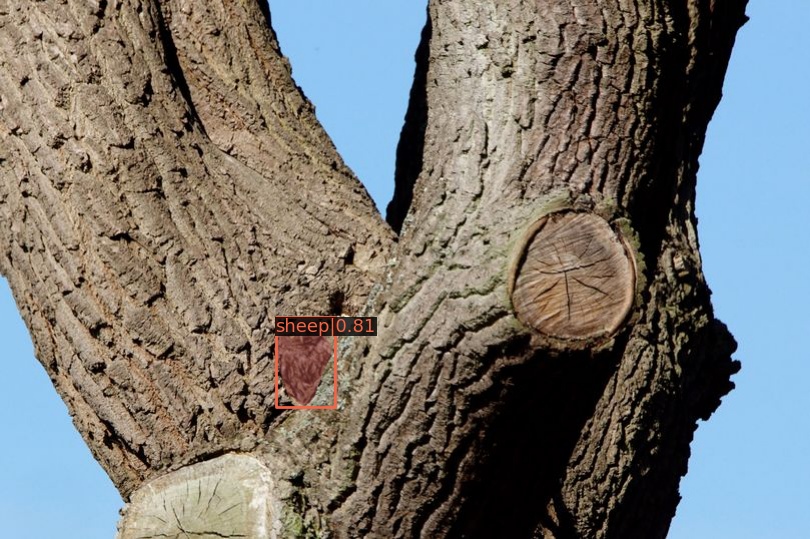} \label{fig:human-logical-amodal-0} \hfill
    \includegraphics[height = 1.25cm]{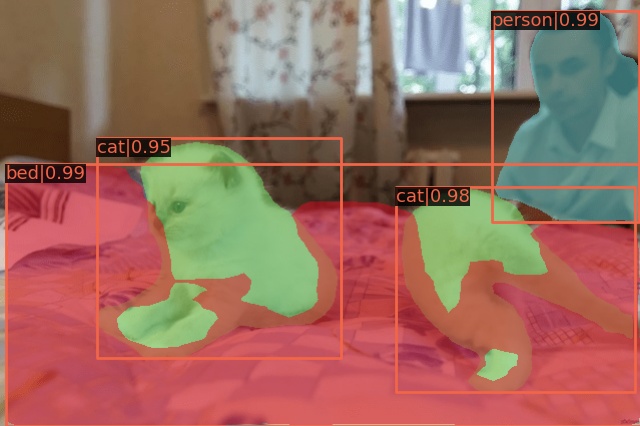} \label{fig:human-logical-amodal-0} \hfill    
    
    \subfloat[ \label{fig:human-logical-cat-fg}]{\includegraphics[height = 1.25cm]{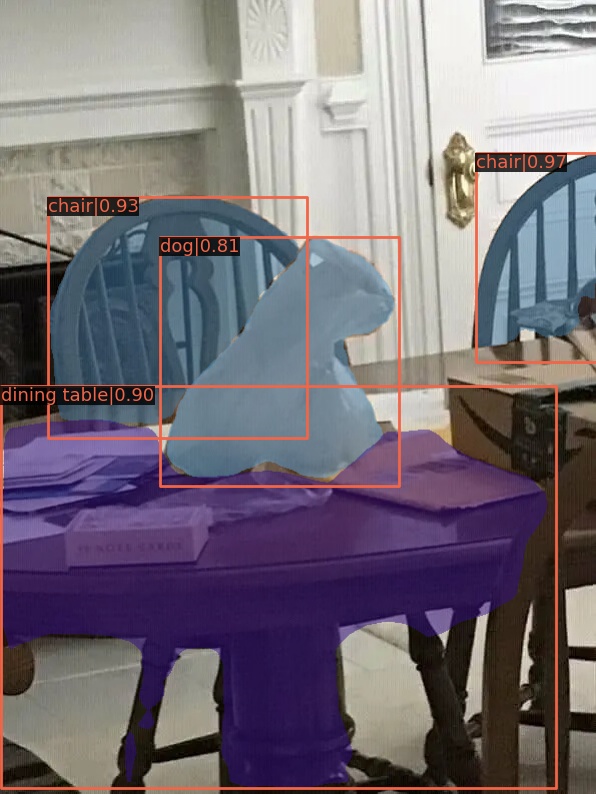}}   \hfill   
    \subfloat[ \label{fig:human-logical-tree-fg}]{\includegraphics[height = 1.25cm]{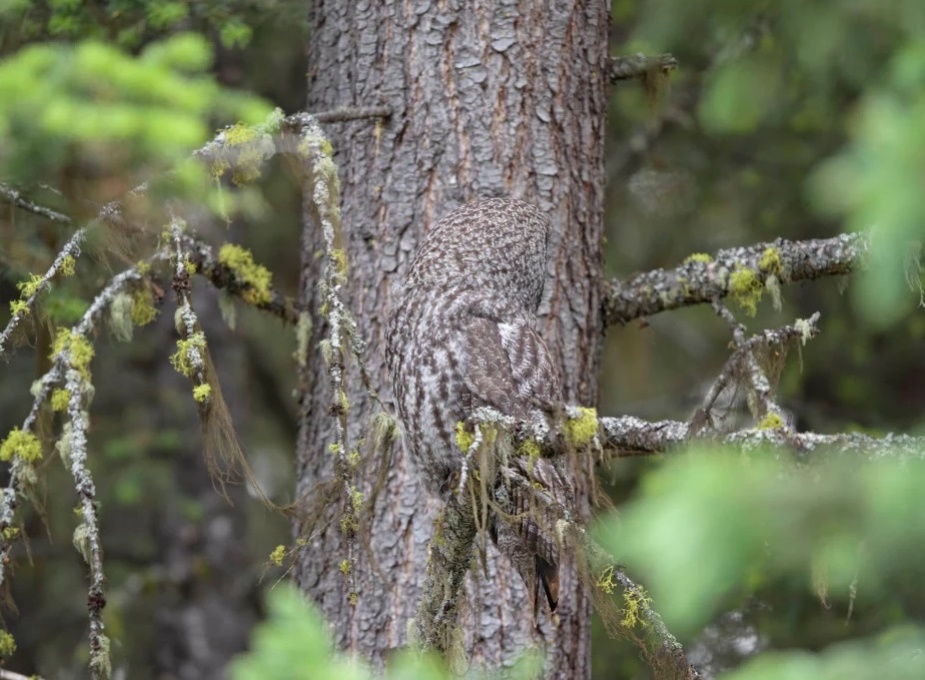}}   \hfill  
    \subfloat[ \label{fig:human-logical-impossible}]{\includegraphics[height = 1.25cm]{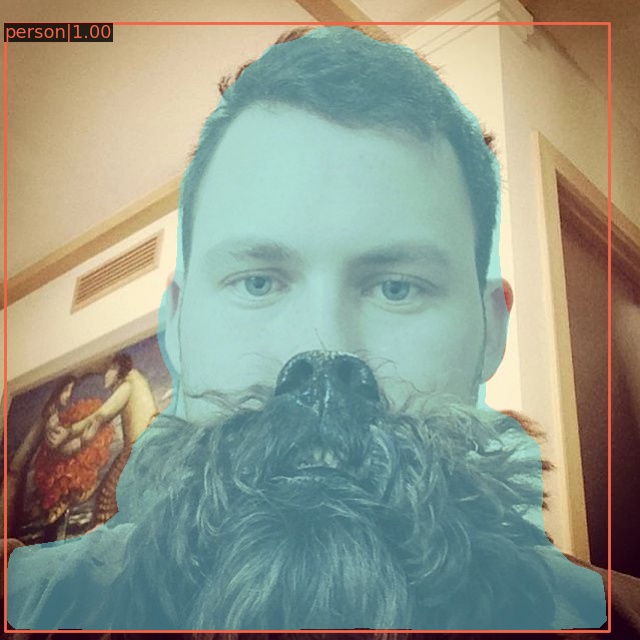}}   \hfill    
    \subfloat[ \label{fig:human-logical-camo-bird}]{\includegraphics[height = 1.25cm]{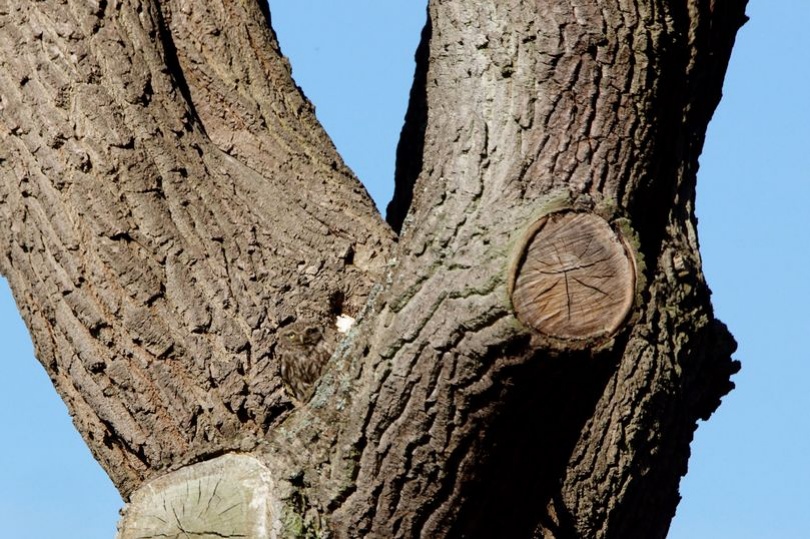}}   \hfill   
    \subfloat[ \label{fig:human-logical-amodal-gaps}]{\includegraphics[height = 1.25cm]{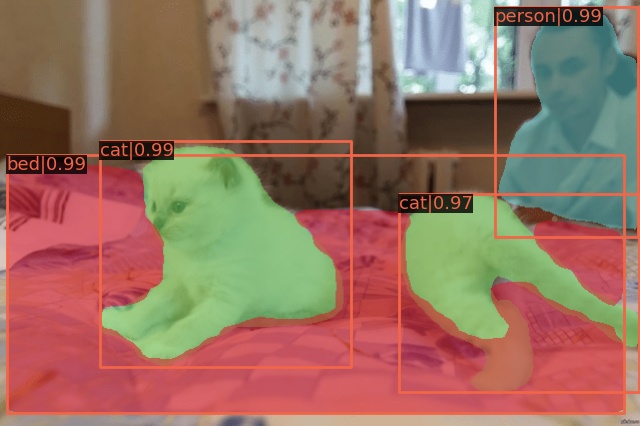}}   \hfill         
    
\caption{ Top row has the input image, second row contains response of Instaboost with ResNet-101 backbone \cite{fang2019instaboost} and third row contain outputs of Swin-Transformer (a) Shape focus in human vision does not text for texture and identifies as cat which is a cover (b,c) Texture focus help models identify the camouflaged bird and the dog (d) The texture focus identifies the bird but assigns wrong class as sheep (e) Amodal closure and the second row shows absence of sub-component formation causing errors on boundaries.}
\label{fig:short-cuts}
\end{figure}

Our study observes that human vision uses a top-down approach similar to Gestalt theory, focusing on overall structure and shape, compared to deep learning models that focus more on details, similar to the structuralist theory of human vision. In many cases, the models generate outputs with both correct perception and perception similar to intuitive processing, as they do not learn the constraints in eq.\ref{eq:non-exclusive}, a trivial logic in human visual processing. In general, the texture bias in deep learning models helped segmentation models perform slightly better on figure-ground errors ($24-40\%$) (e.g., Fig.\ref{fig:human-logical-cat-fg}, \ref{fig:human-logical-tree-fg} and \ref{fig:box-correct}), and the smaller receptive field helps segmentation models avoid amodal closure errors in our adversarial dataset ($23-66\%$) (Fig.\ref{fig:box-correct}). The dataset and the proposed approach can help evaluate changes in the qualitative behavior of models with changes in model architecture, training procedure, or dataset and augmentation approaches. Despite better performance in categories like amodal closure and figure-ground errors, all the segmentation models predicted wrongly on $47.81\%$ of images, and swin-transformer was most accurate at $34\%$. 

\begin{figure*}[t]
\centering        
    \begin{minipage}{0.2\linewidth}
    \includegraphics[height=0.8\linewidth]{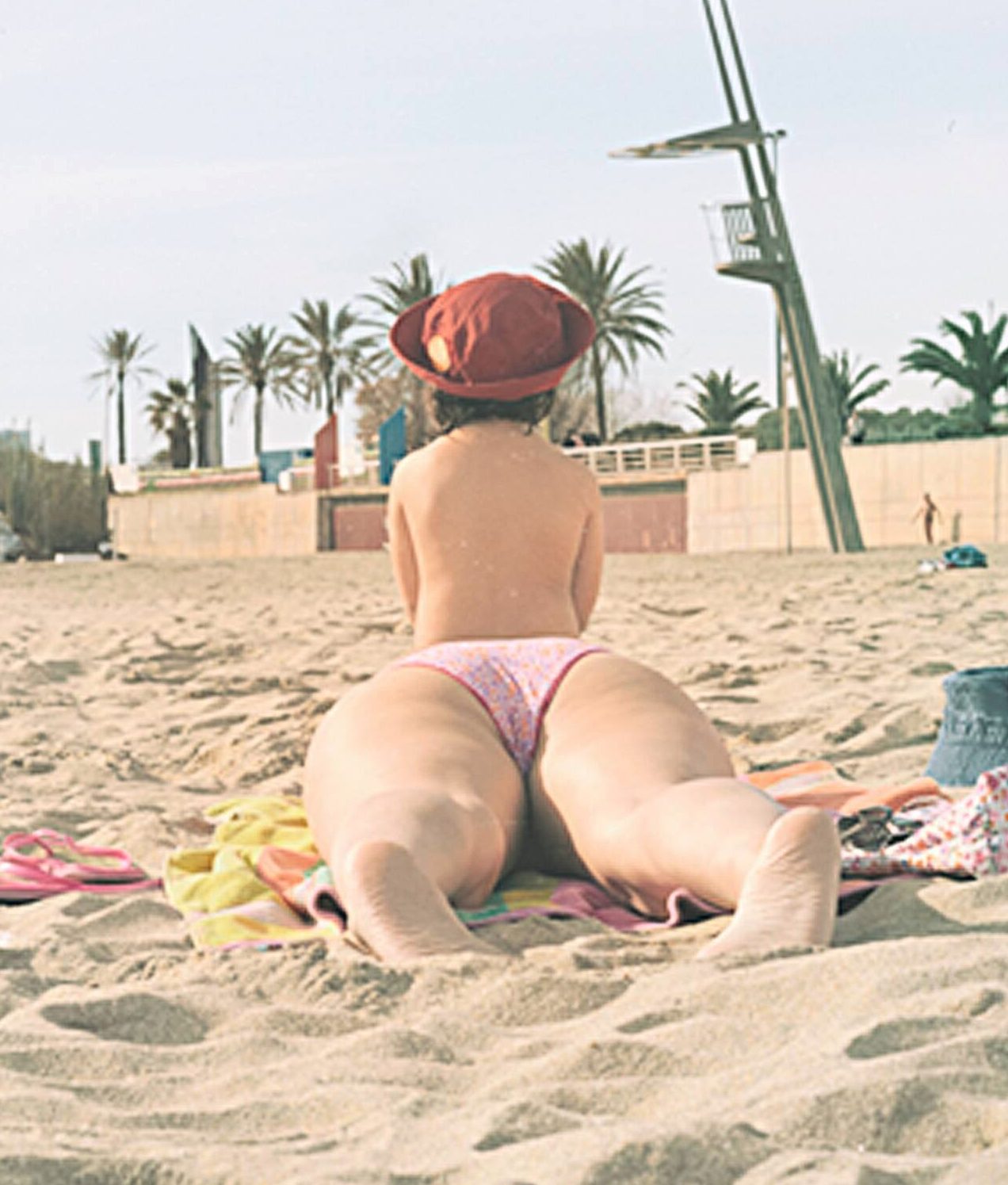}    
    \includegraphics[height=0.8\linewidth]{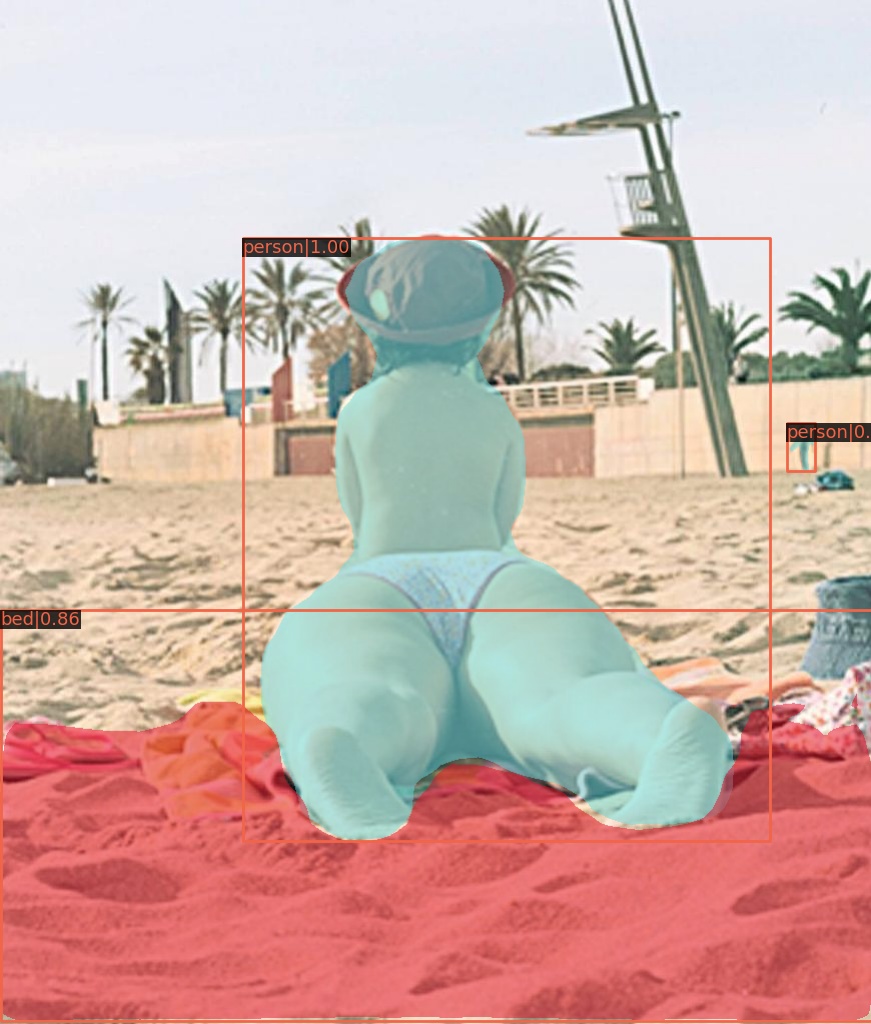}    
    \caption*{\textit{\textbf{How many people can you see in the centre of the picture ?}}\\
        \textit{\textbf{\textcolor{gray}{o3 : } \textbf{\textcolor{red}{1}} }}\\
        \textit{\textbf{\textcolor{gray}{gpt-4o : } \textbf{\textcolor{red}{1}} }}\\
        \textit{\textbf{\textcolor{gray}{gpt-4o-mini : } \textbf{\textcolor{red}{1}} }}\\
        \textit{\textbf{\textcolor{gray}{llama-3.2-11B : }} \textbf{\textcolor{red}{1}} }}
        \textit{\textbf{\textcolor{gray}{llama-3.2-90B : } \textbf{\textcolor{red}{1}} }}\\
        \textit{\textbf{\textcolor{gray}{Qwen2.5-VL-7B-Instruct} \textbf{\textcolor{red}{1}} }}\\
        \textit{\textbf{\textcolor{gray}{Qwen2.5-VL-72B-Instruct : } \textbf{\textcolor{red}{1}} }}\\
        \textit{\textbf{\textcolor{gray}{Pixtral-12B : } \textbf{\textcolor{red}{1}} }}\\
        \textit{\textbf{\textcolor{gray}{gemini-2.0-flash : } \textbf{\textcolor{red}{1}} }}\\
        \textit{\textbf{\textcolor{gray}{deepseek-vl2 : } \textbf{\textcolor{red}{1}} }}\\
        \textit{\textbf{\textcolor{gray}{sonnet-4 : } \textbf{\textcolor{red}{1}} }}
    \caption*{(a) Size Difference}    
    \end{minipage}  \hfill \begin{minipage}{0.2\textwidth}\end{minipage}
    \begin{minipage}{0.18\linewidth}
    \includegraphics[height=0.85\linewidth]{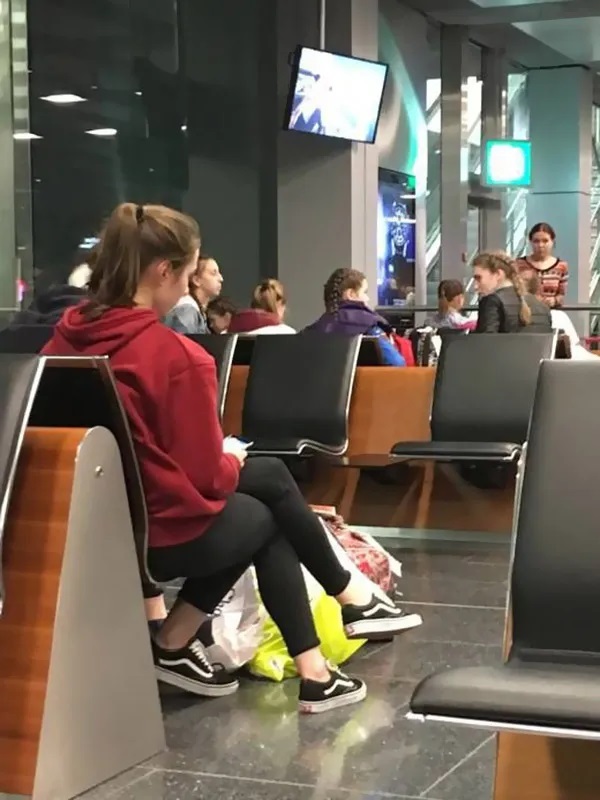}    
    \includegraphics[height=0.85\linewidth]{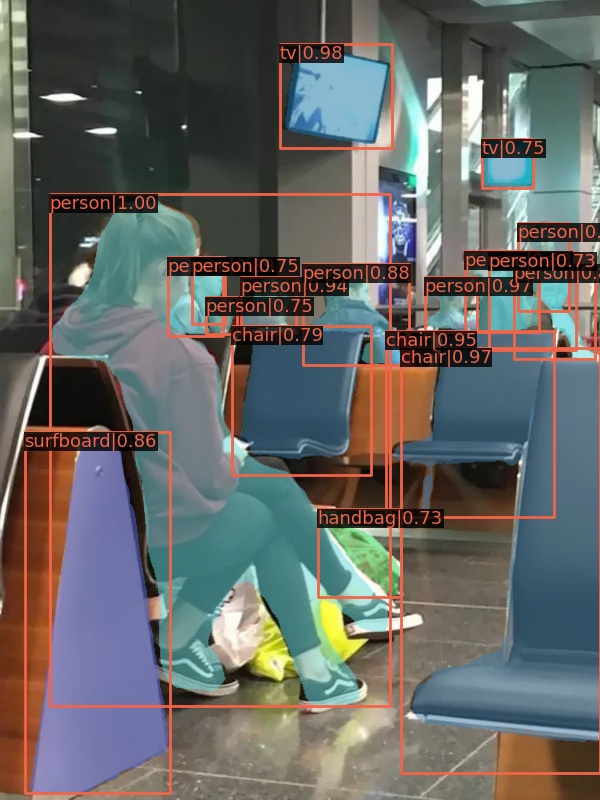}    
    \caption*{\textit{\textbf{How many legs of the woman are visible in this picture ?}}\\
        \textit{\textbf{\textit{\textbf{\textcolor{gray}{o3 : } \textbf{\textcolor{green}{2}} }}\\
        \textcolor{gray}{gpt-4o : } \textbf{\textcolor{green}{2}} }}\\
        \textit{\textbf{\textcolor{gray}{gpt-4o-mini : } \textbf{\textcolor{green}{2}} }}\\
        \textit{\textbf{\textcolor{gray}{llama-3.2-11B : }} \textbf{\textcolor{red}{4}} }}
         \textit{\textbf{\textcolor{gray}{llama-3.2-90B : } \textbf{\textcolor{green}{2}} }}\\
        \textit{\textbf{\textcolor{gray}{Qwen2.5-VL-7B-Instruct} \textbf{\textcolor{red}{4}} }}\\
        \textit{\textbf{\textcolor{gray}{Qwen2.5-VL-72B-Instruct : } \textbf{\textcolor{red}{4}} }}\\
        \textit{\textbf{\textcolor{gray}{Pixtral-12B : } \textbf{\textcolor{green}{2}} }}\\
        \textit{\textbf{\textcolor{gray}{gemini-2.0-flash : } \textbf{\textcolor{green}{2}} }}\\
        \textit{\textbf{\textcolor{gray}{deepseek-vl2 : } \textbf{\textcolor{green}{2}} }}\\
        \textit{\textbf{\textcolor{gray}{sonnet-4 : } \textbf{\textcolor{green}{2}} }}
    \caption*{(b) Count Difference}    
    \end{minipage} \hfill \begin{minipage}{0.2\textwidth}\end{minipage}
    \begin{minipage}{0.2\linewidth} 
    \includegraphics[height=0.8\linewidth]{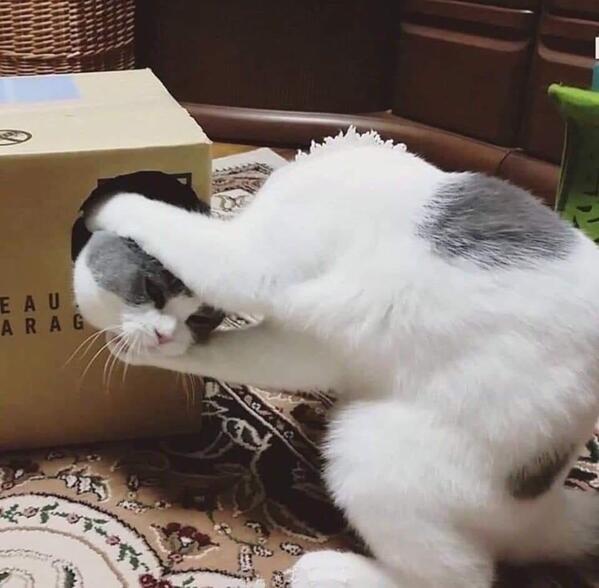}    
    \includegraphics[height=0.8\linewidth]{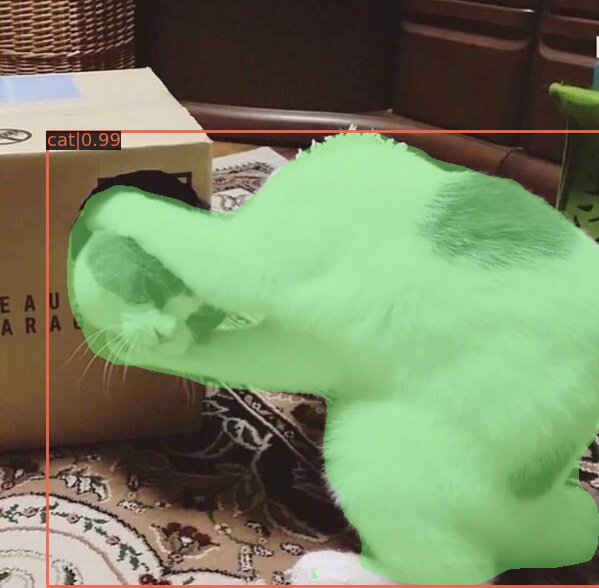}    
    \caption*{\textit{\textbf{How many cats are there in the picture ? \\ }}\\
        \textit{\textit{\textbf{\textcolor{gray}{o3 : } \textbf{\textcolor{red}{1}} }}\\
        \textbf{\textcolor{gray}{gpt-4o : } \textbf{\textcolor{red}{1}} }}\\
        \textit{\textbf{\textcolor{gray}{gpt-4o-mini : } \textbf{\textcolor{red}{1}} }}\\
        \textit{\textbf{\textcolor{gray}{llama-3.2-11B : }} \textbf{\textcolor{red}{1}} }}
         \textit{\textbf{\textcolor{gray}{llama-3.2-90B : } \textbf{\textcolor{red}{1}} }}\\
        \textit{\textbf{\textcolor{gray}{Qwen2.5-VL-7B-Instruct} \textbf{\textcolor{red}{1}} }}\\
        \textit{\textbf{\textcolor{gray}{Qwen2.5-VL-72B-Instruct : } \textbf{\textcolor{red}{1}} }}\\
        \textit{\textbf{\textcolor{gray}{Pixtral-12B : } \textbf{\textcolor{red}{1}} }}\\
        \textit{\textbf{\textcolor{gray}{gemini-2.0-flash : } \textbf{\textcolor{red}{1}} }}\\
        \textit{\textbf{\textcolor{gray}{deepseek-vl2 : } \textbf{\textcolor{red}{1}} }}\\
        \textit{\textbf{\textcolor{gray}{sonnet-4 : } \textbf{\textcolor{red}{1}} }}
    \caption*{(c) Impossible Posture}    
    \end{minipage}  \hfill \begin{minipage}{0.2\textwidth}\end{minipage}
    \begin{minipage}{0.2\linewidth}
    \includegraphics[height=0.8\linewidth]{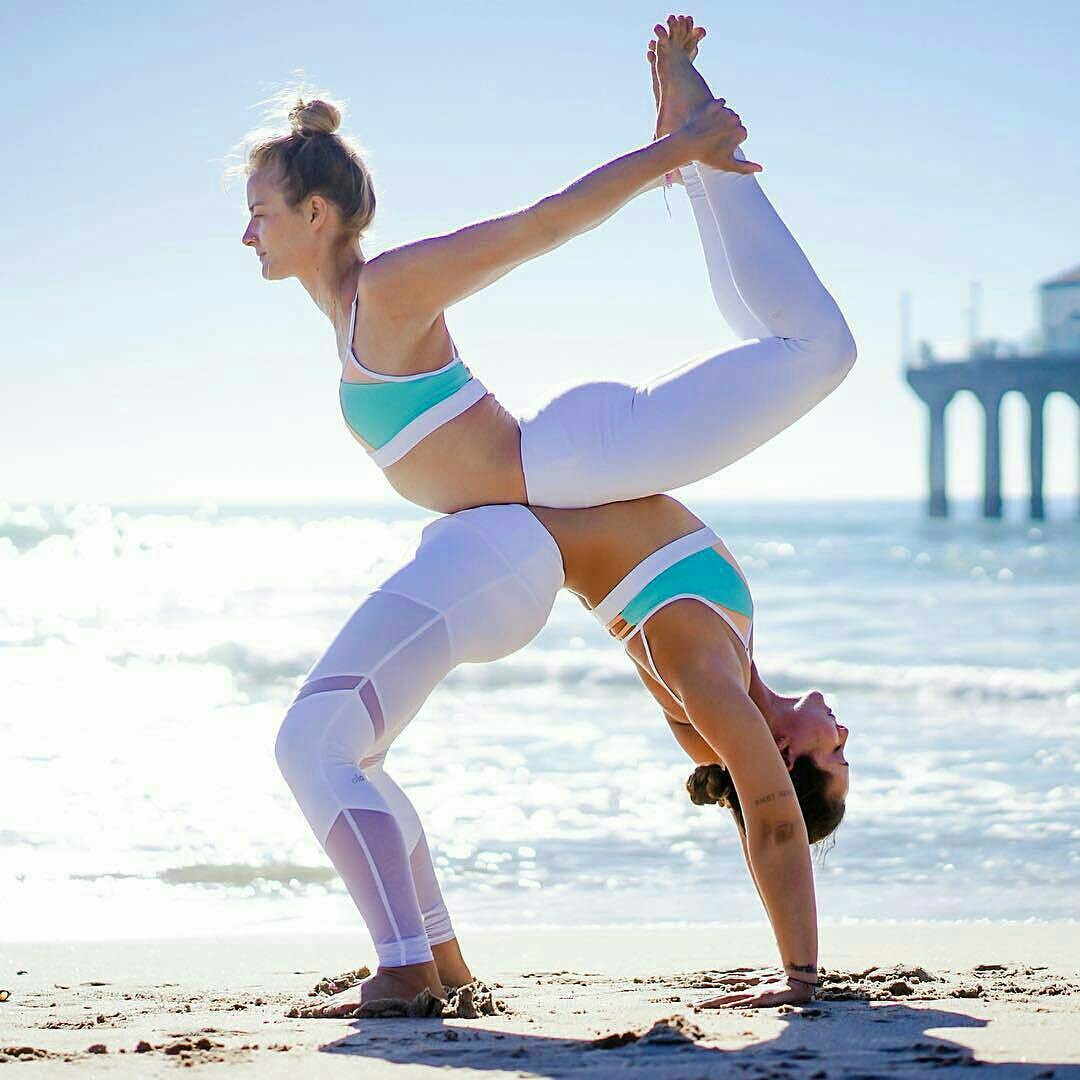}    
    \includegraphics[height=0.8\linewidth]{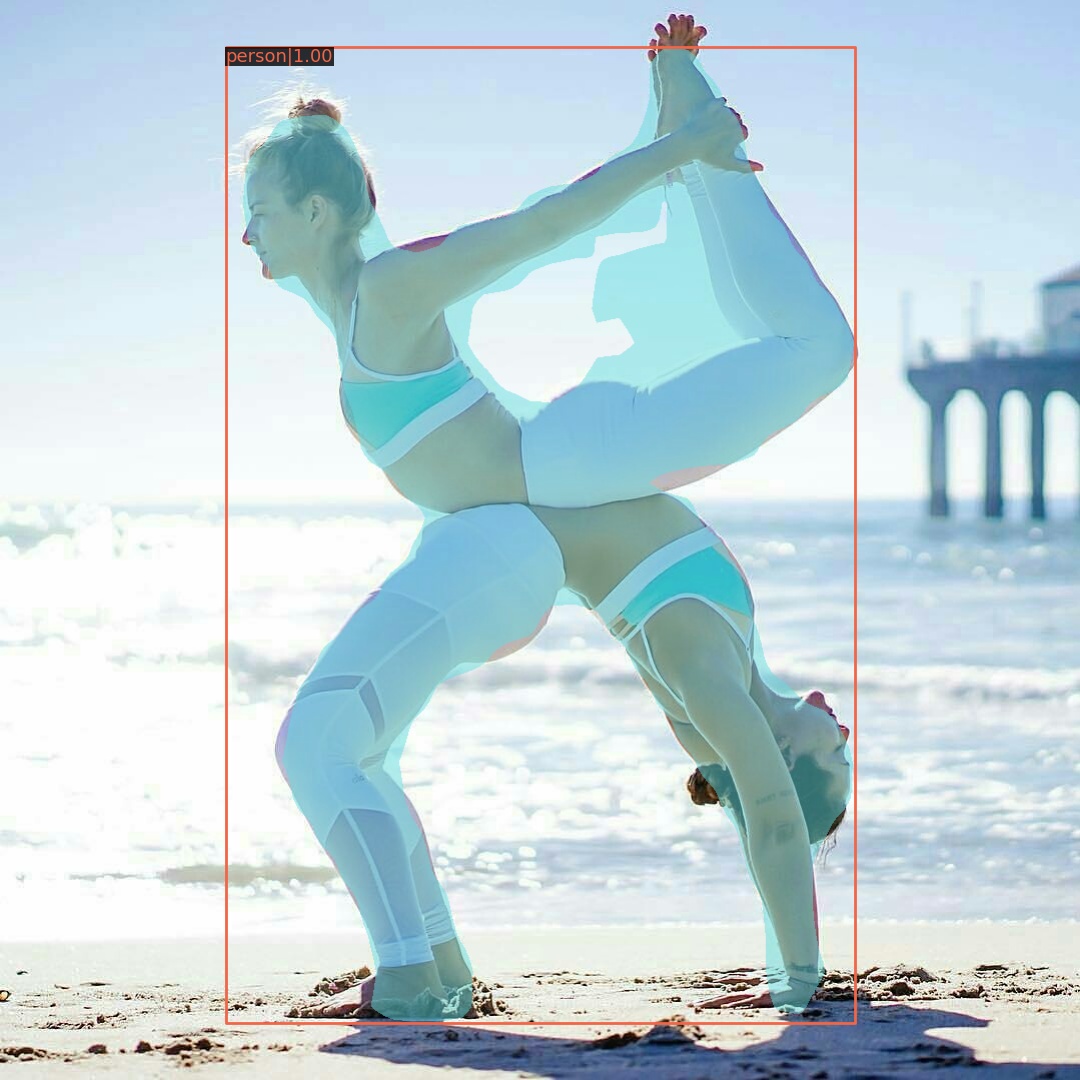}    
    \caption*{\textit{\textbf{Is the lady standing ? \\ \\ }}\\
        \textit{\textbf{\textit{\textbf{\textcolor{gray}{o3 : } \textbf{\textcolor{green}{No}} }}\\
        \textcolor{gray}{gpt-4o : } \textbf{\textcolor{green}{No}} }}\\
        \textit{\textbf{\textcolor{gray}{gpt-4o-mini : } \textbf{\textcolor{green}{No}} }}\\
        \textit{\textbf{\textcolor{gray}{llama-3.2-11B : }} \textbf{\textcolor{green}{No}} }} \textit{\textbf{\textcolor{gray}{llama-3.2-90B : } \textbf{\textcolor{green}{No}} }}\\
        \textit{\textbf{\textcolor{gray}{Qwen2.5-VL-7B-Instruct} \textbf{\textcolor{green}{No   }} }}\\
        \textit{\textbf{\textcolor{gray}{Qwen2.5-VL-72B-Instruct : } \textbf{\textcolor{green}{No}} }}\\
        \textit{\textbf{\textcolor{gray}{Pixtral-12B : } \textbf{\textcolor{red}{Yes}} }}\\
        \textit{\textbf{\textcolor{gray}{gemini-2.0-flash : } \textbf{\textcolor{green}{No}} }}\\
        \textit{\textbf{\textcolor{gray}{deepseek-vl2 : } \textbf{\textcolor{green}{No}} }}\\
        \textit{\textbf{\textcolor{gray}{sonnet-4 : } \textbf{\textcolor{green}{No}} }}
    \caption*{(d) Impossible Posture}    
    
    \end{minipage}    
\caption{Input image, segmentation model (swin-t) output and response of multi-modal LLMs. Segmentation models shows poor performance on logical errors, where as VLMs have made progress. Size differences are difficult for VLMs still to understand.}
\label{fig:commonLogicalErrors-all}
\end{figure*}

YOLACT model \cite{bolya2019yolact}, developed for real-time processing, utilizes an architecture that generates and groups sub-components. YOLACT demonstrated a shape bias of 17, significantly higher than the next highest model, GroIE \cite{rossi2021novel}, which exhibited a shape bias of only 5.5 on images in which the global shape aligned with one class but comprised two distinct instances from different classes (e.g., Fig. \ref{fig:human-dog-mix}). In these instances, intuitive processing predominantly relies on shape, whereas accurate perception requires attention to texture. YOLACT’s pronounced shape bias suggests that, like human vision, segmentation models inherently learn to use shape cues for grouping sub-components. Additionally, YOLACT exhibited a higher ratio of human intuitive references to correct inferences overall and across Gestalt properties, except for amodal closure errors. Future research can further investigate segmentation models as computational frameworks for modeling intuitive processing in human vision and address a limitation of using classification models, which lack object localization capabilities.

\input{sec/6_llms}

\subsection{Logical Processing}

We examine logical processing in human vision, which helps detect and correct errors that arise in instances formed during intuitive processing. Unlike earlier errors that become apparent through finer details, these errors require deeper evaluation. We explore how human vision assesses the validity, integrity, and coherence of instances perceived in intuitive processing. Using thematic analysis, we identify common logical errors in perception, analyze the mechanisms of logical processing in vision, and compare the ability of segmentation and vision-language models to interpret such images. The mistakes in intuitive processing in human vision are corrected during subsequent processing, whereas segmentation models face a significant limitation in their inability to identify and rectify logical errors. The segmentation models lack sub-component structure and depend on probabilistic grouping of components, leading to errors related to count mismatches, size differences, and failure to recognize impossible patterns.

\begin{figure*}[ht]
\centering      

    \begin{minipage}{0.18\linewidth}
    \centering
    \includegraphics[height=0.85\linewidth]{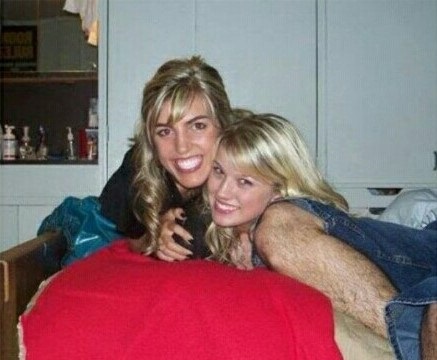}    
    \caption*{\textit{\textbf{How many people are there in this picture ?}}\\
        \textit{\textbf{\textcolor{gray}{gpt-4o : } \textbf{\textcolor{red}{2}} }}\\
        \textit{\textbf{\textcolor{gray}{gpt-4o-mini : } \textbf{\textcolor{red}{2}} }}\\
        \textit{\textbf{\textcolor{gray}{llama-3.2-11B : }} \textbf{\textcolor{red}{2}} }}
    \end{minipage} \hfill \begin{minipage}{0.2\textwidth}\end{minipage}
    \begin{minipage}{0.18\linewidth}
    \centering
    \includegraphics[height=0.85\linewidth]{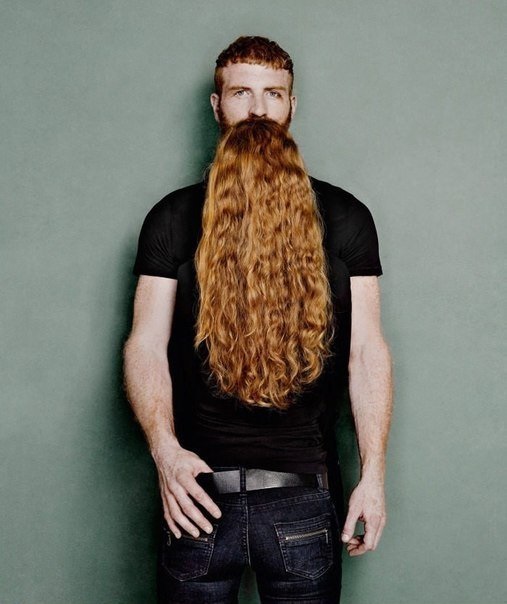}    
    \caption*{\textit{\textbf{How many people are there in the picture?}}\\
        \textit{\textbf{\textcolor{gray}{gpt-4o : } \textbf{\textcolor{red}{1}} }}\\
        \textit{\textbf{\textcolor{gray}{gpt-4o-mini : } \textbf{\textcolor{red}{1}} }}\\
        \textit{\textbf{\textcolor{gray}{llama-3.2-11B : }} \textbf{\textcolor{red}{1}} }}
    \end{minipage} \hfill \begin{minipage}{0.2\textwidth}\end{minipage}
    \begin{minipage}{0.18\linewidth}
    \centering
    \includegraphics[height=0.85\linewidth]{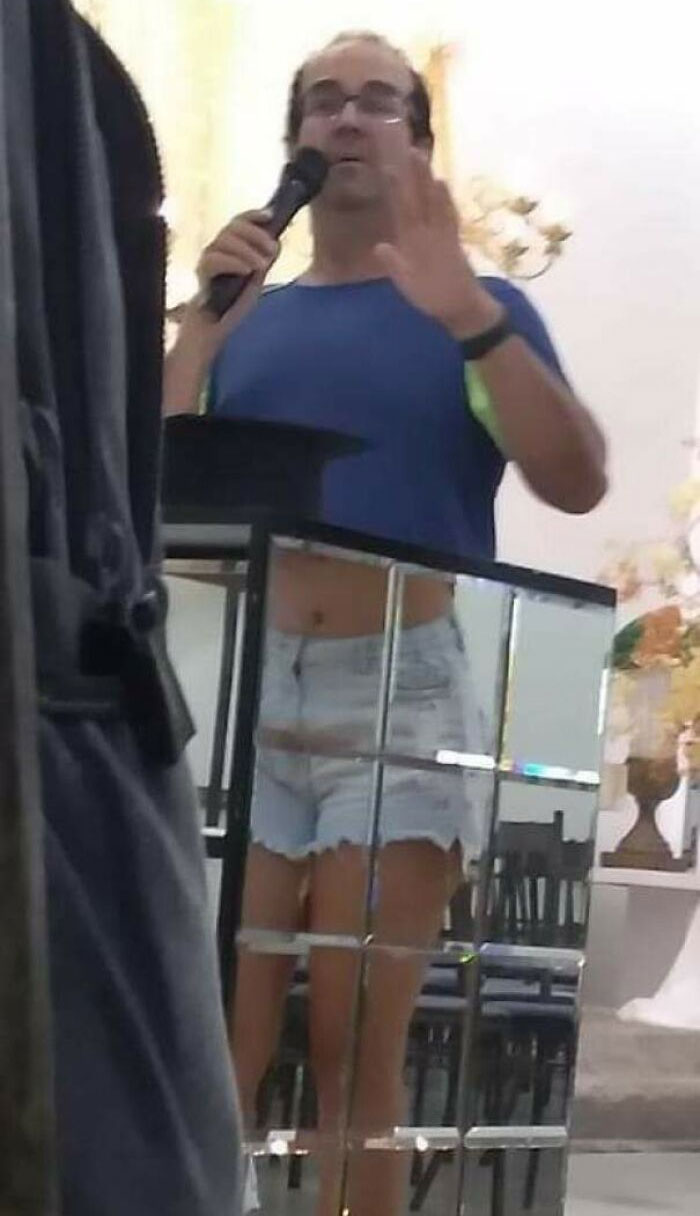}    
    \caption*{\textit{\textbf{Can you see a man wearing jean shorts?}}\\
        \textit{\textbf{\textcolor{gray}{gpt-4o : } \textbf{\textcolor{red}{Yes.}} }}\\
        \textit{\textbf{\textcolor{gray}{gpt-4o-mini : } \textbf{\textcolor{red}{Yes}} }}\\
        \textit{\textbf{\textcolor{gray}{llama-3.2-11B : }} \textbf{\textcolor{red}{Yes.}} }}
    \end{minipage} \hfill \begin{minipage}{0.2\textwidth}\end{minipage}
    \begin{minipage}{0.18\linewidth}
    \centering
    \includegraphics[height=0.85\linewidth]{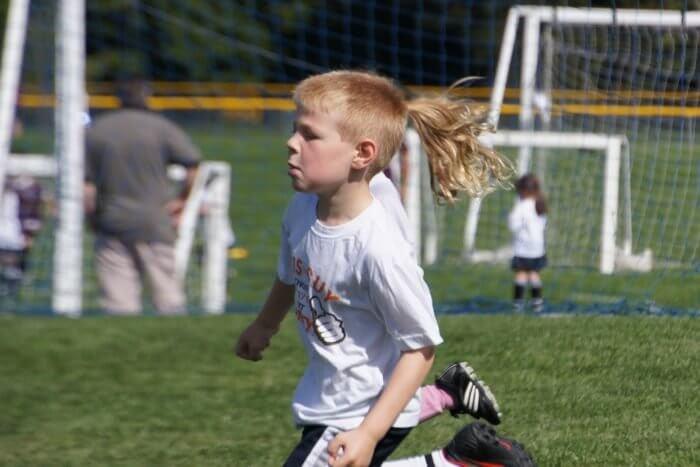}    
    \caption*{\textit{\textbf{Does the boy have a ponytail ?}}\\
        \textit{\textbf{\textcolor{gray}{gpt-4o : } \textbf{\textcolor{red}{Yes.}} }}\\
        \textit{\textbf{\textcolor{gray}{gpt-4o-mini : } \textbf{\textcolor{red}{Yes}} }}\\
        \textit{\textbf{\textcolor{gray}{llama-3.2-11B : }} \textbf{\textcolor{red}{Yes.}} }}
    \end{minipage} \hfill \begin{minipage}{0.2\textwidth}\end{minipage}
\caption{They lack finer reasoning of sub-components as in human vision (e.g., wrongly merged components of different genders are identified from clothing and accessories), and such examples requiring finer analysis may be rare in the dataset, making it difficult to learn during training \cite{francazi2023theoretical}. Examples of images requiring fine analysis of patterns for correct perception.}
\label{fig:llm-finer}
\end{figure*}
\begin{figure*}[ht]
\centering      
    \begin{minipage}{0.18\linewidth}
    \centering
    \includegraphics[height=0.85\linewidth]{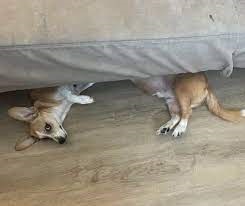}    
    \caption*{\textit{\textbf{how many dogs are there in the picture ?}}\\
        \textit{\textbf{\textcolor{gray}{gpt-4o : } \textbf{\textcolor{red}{1}} }}\\
        \textit{\textbf{\textcolor{gray}{gpt-4o-mini : } \textbf{\textcolor{red}{1}} }}\\
        \textit{\textbf{\textcolor{gray}{llama-3.2-11B : }} \textbf{\textcolor{red}{1}} }}
    \end{minipage} \hfill \begin{minipage}{0.2\textwidth}\end{minipage}    
    \begin{minipage}{0.18\linewidth}
    \centering
    \includegraphics[height=0.85\linewidth]{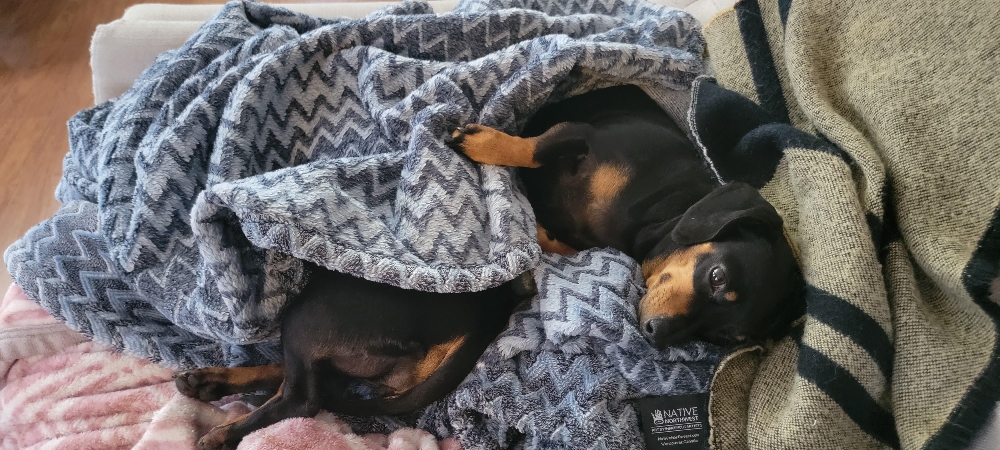}    
    \caption*{\textit{\textbf{How many dogs are in this picture ?}}\\
        \textit{\textbf{\textcolor{gray}{gpt-4o : } \textbf{\textcolor{red}{1}} }}\\
        \textit{\textbf{\textcolor{gray}{gpt-4o-mini : } \textbf{\textcolor{red}{1}} }}\\
        \textit{\textbf{\textcolor{gray}{llama-3.2-11B : }} \textbf{\textcolor{red}{1}} }}
    \end{minipage} \hfill \begin{minipage}{0.2\textwidth}\end{minipage}    
    \begin{minipage}{0.18\linewidth}
    \centering
    \includegraphics[height=0.85\linewidth]{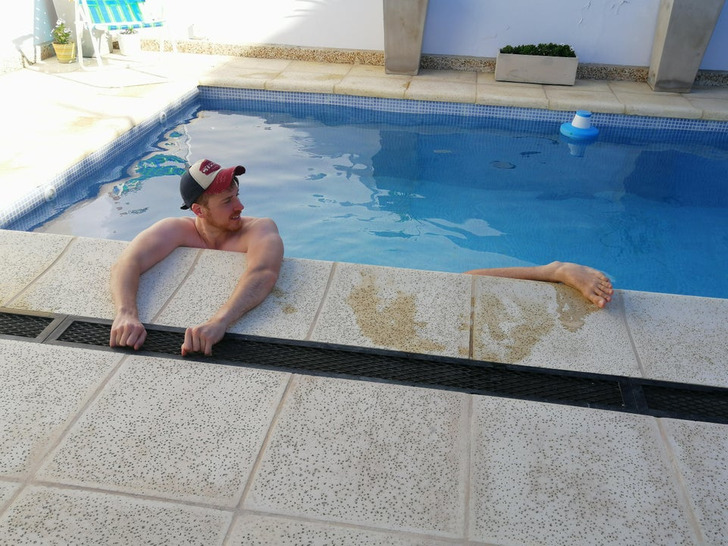}    
    \caption*{\textit{\textbf{How many people do you see in the picture}}\\
        \textit{\textbf{\textcolor{gray}{gpt-4o : } \textbf{\textcolor{red}{1}} }}\\
        \textit{\textbf{\textcolor{gray}{gpt-4o-mini : } \textbf{\textcolor{red}{1}} }}\\
        \textit{\textbf{\textcolor{gray}{llama-3.2-11B : }} \textbf{\textcolor{red}{1}} }}
    \end{minipage} \hfill \begin{minipage}{0.2\textwidth}\end{minipage}
    \begin{minipage}{0.18\linewidth}
    \centering
    \includegraphics[height=0.85\linewidth]{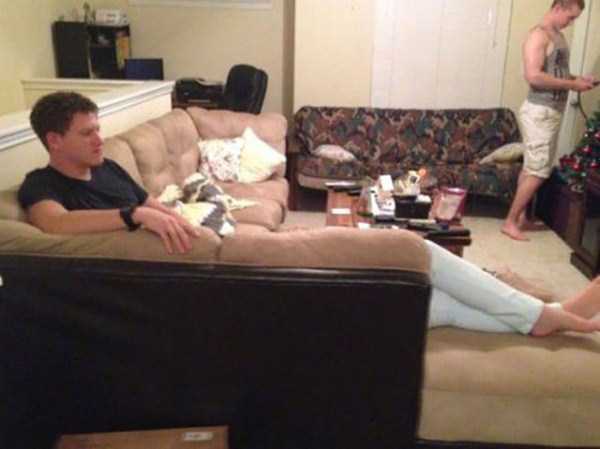}    
    \caption*{\textit{\textbf{How many people are in the picture ?}}\\
        \textit{\textbf{\textcolor{gray}{gpt-4o : } \textbf{\textcolor{red}{2}} }}\\
        \textit{\textbf{\textcolor{gray}{gpt-4o-mini : } \textbf{\textcolor{red}{2}} }}\\
        \textit{\textbf{\textcolor{gray}{llama-3.2-11B : }} \textbf{\textcolor{red}{2}} }}
    \end{minipage} \hfill \begin{minipage}{0.2\textwidth}\end{minipage}
\caption{Multi-modal LLMs also learns amodal closure as an emergent property and tend to merge disconnected components by shape, hence performing poorly on adversarial examples. Examples from amodal closure errors where all the large language models made wrong inference.}
\label{fig:llm-amodal}
\end{figure*}
\begin{figure*}[ht]
\centering      

    \begin{minipage}{0.18\linewidth} %
    \centering
    \includegraphics[height=0.85\linewidth]{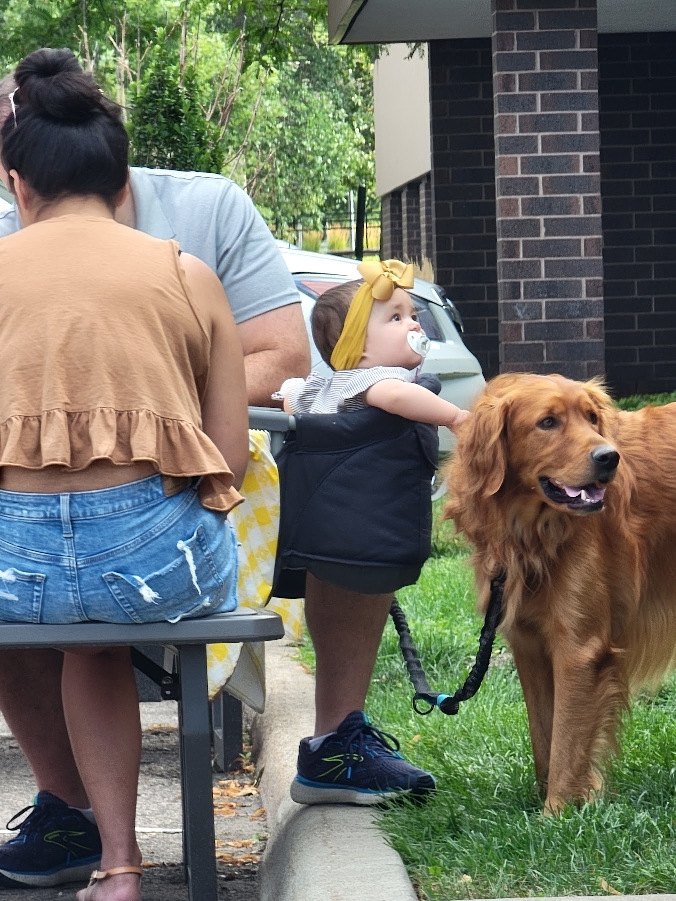}    
    \caption*{\textit{\textbf{Can you see the child's leg ?}}\\
        \textit{\textbf{\textcolor{gray}{gpt-4o : } \textbf{\textcolor{red}{Yes.}} }}\\
        \textit{\textbf{\textcolor{gray}{gpt-4o-mini : } \textbf{\textcolor{red}{Yes.}} }}\\
        \textit{\textbf{\textcolor{gray}{llama-3.2-11B : }} \textbf{\textcolor{red}{Yes.}} }}
    \end{minipage} \hfill \begin{minipage}{0.2\textwidth}\end{minipage}
    \begin{minipage}{0.18\linewidth} %
    \centering
    \includegraphics[height=0.85\linewidth]{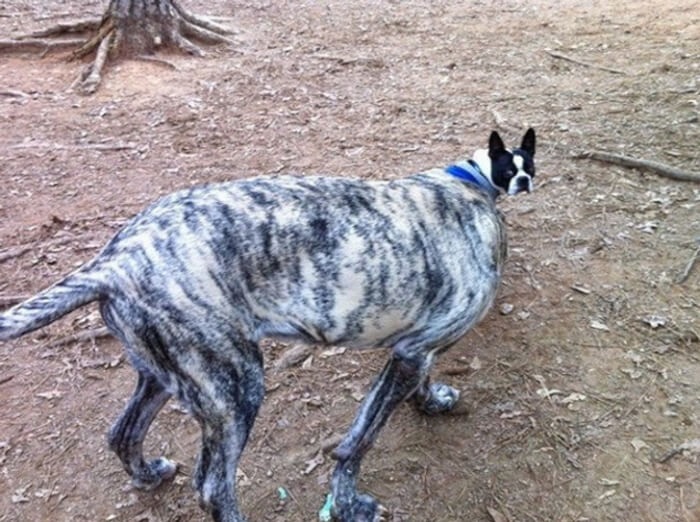}    
    \caption*{\textit{\textbf{How many dogs are in this picture ?}}\\
        \textit{\textbf{\textcolor{gray}{gpt-4o : } \textbf{\textcolor{red}{1}} }}\\
        \textit{\textbf{\textcolor{gray}{gpt-4o-mini : } \textbf{\textcolor{red}{1}} }}\\
        \textit{\textbf{\textcolor{gray}{llama-3.2-11B : }} \textbf{\textcolor{red}{1}} }}
    \end{minipage} \hfill \begin{minipage}{0.2\textwidth}\end{minipage}
    \begin{minipage}{0.18\linewidth} %
    \centering
    \includegraphics[height=0.85\linewidth]{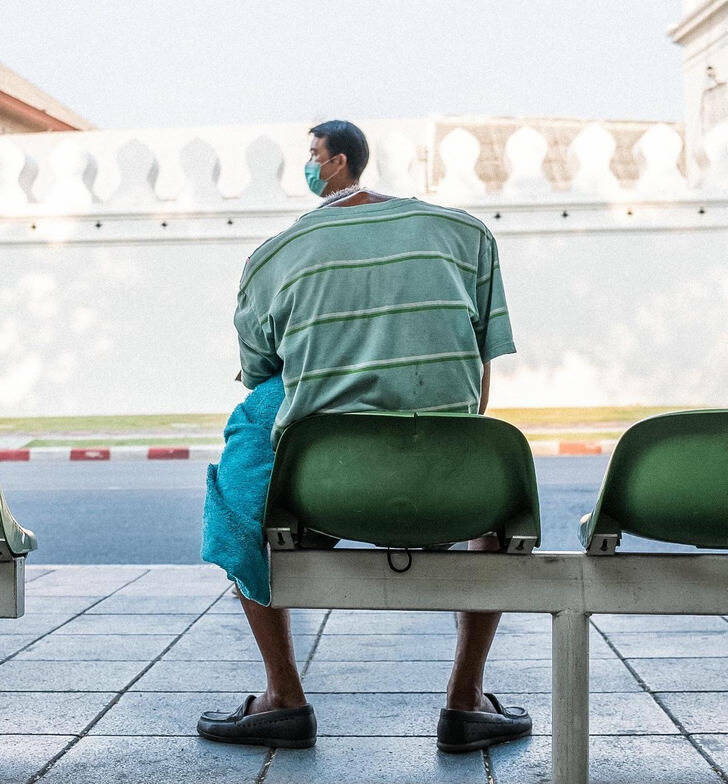}    
    \caption*{\textit{\textbf{How many people are there in this picture ?}}\\
        \textit{\textbf{\textcolor{gray}{gpt-4o : } \textbf{\textcolor{red}{1}} }}\\
        \textit{\textbf{\textcolor{gray}{gpt-4o-mini : } \textbf{\textcolor{red}{1}} }}\\
        \textit{\textbf{\textcolor{gray}{llama-3.2-11B : }} \textbf{\textcolor{red}{1}} }}
    \end{minipage} \hfill \begin{minipage}{0.2\textwidth}\end{minipage}
    \begin{minipage}{0.18\linewidth} %
    \centering
    \includegraphics[height=0.85\linewidth]{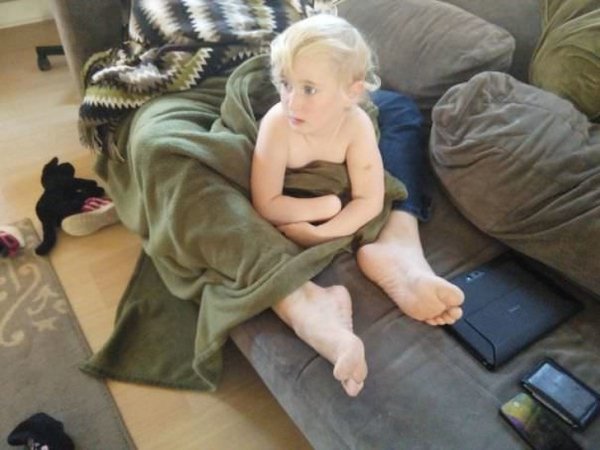}    
    \caption*{\textit{\textbf{Can you see the kid's leg ?}}\\
        \textit{\textbf{\textcolor{gray}{gpt-4o : } \textbf{\textcolor{red}{Yes.}} }}\\
        \textit{\textbf{\textcolor{gray}{gpt-4o-mini : } \textbf{\textcolor{red}{Yes.}} }}\\
        \textit{\textbf{\textcolor{gray}{llama-3.2-11B : }} \textbf{\textcolor{red}{Yes.}} }}
    \end{minipage} \hfill \begin{minipage}{0.2\textwidth}\end{minipage}    
\caption{ Multi-modal LLMs have made progress but are still in the lower ranges ($37-61\%$) on images with size difference in components.}
\label{fig:size-diff-examples}
\end{figure*}

\textbf{\textit{Sub-component Differences :}} 
Human vision can distinguish instances where objects of different sizes appear together with the shape of a single object in the second stage of processing. Using overall shape to group components can make intuitive processing prone to these errors. Multi-modal LLMs have made progress but are still in the lower ranges ($20-63\%$) on images with size difference in components (Fig.\ref{fig:size-diff-examples}a, \ref{fig:prop-comparison}). Segmentation models had lower accuracy ($6-17\%$) and made errors similar to human intuitive processing ($56-81\%$), as they are unable to recognize the relative size of components. Similarly, multi-modal LLMs also learns amodal closure as an emergent property and tend to merge disconnected components by shape, hence performing poorly on adversarial examples (Fig.\ref{fig:llm-amodal}). Fig. \ref{fig:llm-finer} They lack finer reasoning of sub-components as in human vision (e.g., wrongly merged components of different genders are identified from clothing and accessories), and such examples requiring finer analysis may be rare in the dataset, making it difficult to learn during training \cite{francazi2023theoretical}.

\textbf{\textit{Count Mismatch :}} 
Segmentation models made the same mistake as intuitive processing in human vision in $56-62\%$ of images  (Fig.\ref{fig:box-human}). The components grouped using the most probable shape tend to group nearby components without sub-component structure, and most models had poor accuracy ($3-10\%$). For example, in Fig.\ref{fig:commonLogicalErrors-all}d, the common shape is used to form the first instance, and the remaining components form the next instance in human vision. In many cases, we also observe all the components to be grouped in intuitive processing (e.g., Fig.\ref{fig:commonLogicalErrors-all}b). The models seem to rely on a probabilistic grouping of the components without substructures and identify them as a single instance. Multi-modal LLMs made progress at \textcolor{blue}{($49-79\%$)}, where the errors in count difference occur primarily with sub-components higher than possible for the class and are learnable \cite{gupta2023visual}.

\textbf{\textit{Impossible Patterns :}} 
Human vision can identify objects in poses that are not feasible (e.g., Fig.\ref{fig:commonLogicalErrors-all}c,\ref{fig:commonLogicalErrors-all}d). In the case of segmentation models, this capability is missing as it relies on probabilistic grouping. Human vision also assesses the validity of instances formed by grouping sub-components during intuitive processing, facilitating the identification and correction of errors when a sub-component merges with the background—a limitation inherent in human intuitive processing (e.g., a girl's skirt blending with the background would render an instance consisting solely of the upper body invalid). In these instances, the human visual system can infer the missing part by re-evaluating nearby probable regions to validate the object. Additionally, segmentation models split sub-components and partially assign them to two instances by splitting on proximity rather than assigning a sub-component entirely to one instance. The formation of sub-components in human vision can help avoid these errors.

\begin{figure}[ht]
\centering      
\begin{minipage}{0.48\linewidth}
    \centering
    \includegraphics[width=0.4\linewidth]{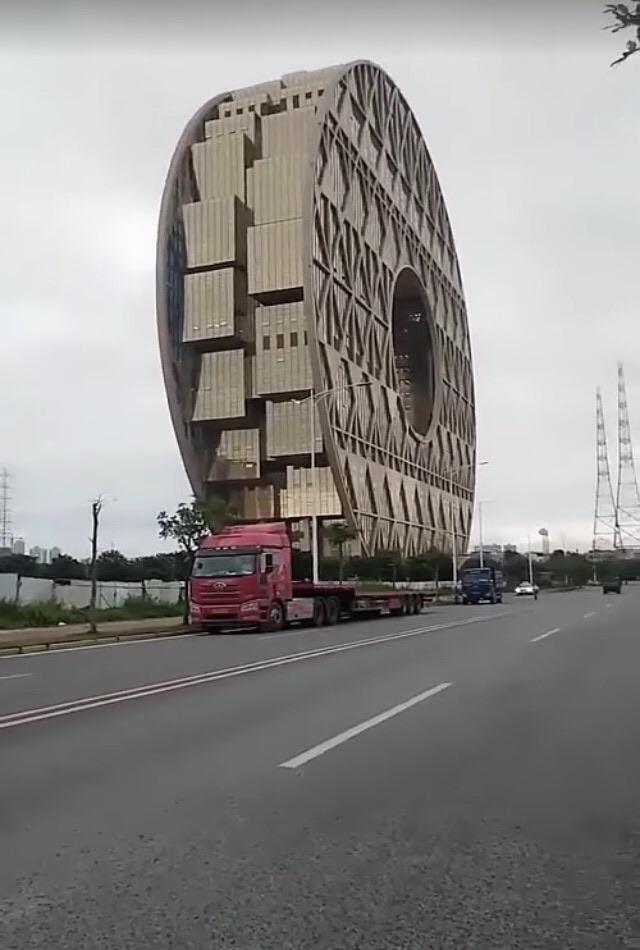}    
    \caption*{\textit{\textbf{Is the truck carrying anything ?}}\\
        \textit{\textbf{\textcolor{gray}{llama-3.2-11B : }} \textbf{\textcolor{red}{The truck is transporting a large, cylindrical structure that appears to be made of metal. The structure has a unique design with a series of rectangular panels and a large circular opening on one side. It is not clear what the purpose of this structure is or what it is used for, but it could be a piece of industrial equipment or a decorative feature.}} }}
\end{minipage} \hfill \begin{minipage}{0.2\textwidth}\end{minipage}
\begin{minipage}{0.48\linewidth}
    \centering
    \includegraphics[width=0.83\linewidth]{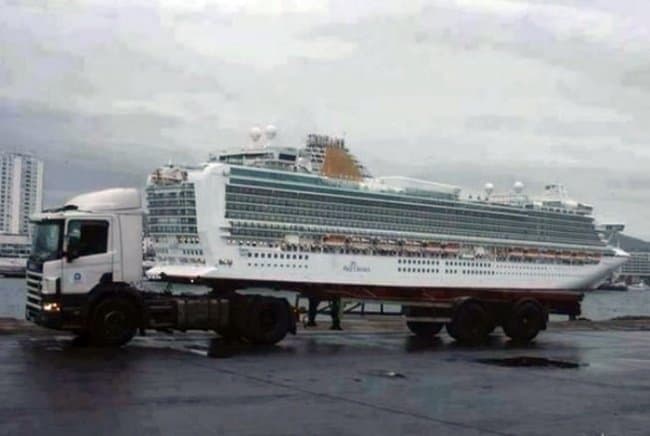}    
    \caption*{\textit{\textbf{Is the truck carrying a cruise ship ?}}\\
        \textit{\textbf{\textcolor{gray}{llama-3.2-11B : }} \textbf{\textcolor{teal}{The truck is not carrying a cruise ship. The image appears to be a joke or a prank, as the truck is not designed to carry a cruise ship and it would be impossible for it to do so. The image is likely meant to be humorous or satirical, rather than a realistic representation of a truck carrying a cruise ship.}} }}
\end{minipage} \hfill \begin{minipage}{0.2\textwidth}\end{minipage}
\caption{Multi-modal LLMs have made progress and show ability to understand substructures and the relationships between sub-components. When queried without constraints limiting responses to Yes/No, these models can identify and localize substructures, likely learning their relationships from textual descriptions rather than explicit annotations.}
\label{fig:llm-substructure}
\end{figure}

Multi-modal LLMs have made progress and show ability to understand substructures and the relationships between sub-components. When queried without constraints limiting responses to Yes/No, these models can identify and localize substructures (Fig.\ref{fig:llm-substructure}), likely learning their relationships from textual descriptions rather than explicit annotations. In contrast, segmentation models lack sub-component structure and rely on grouping sub-components without logical underpinnings. Their poor performance with errors similar to intuitive processing indicates the absence of the second stage of processing that helps detect and recover from these errors (More examples are present in the supplementary data).

In general, models should be correct, robust, and generalizable and not limited by human performance or similarity in strategies to human vision. Deep learning models have been criticized for shortcut learning by biasing on texture \cite{geirhos2020shortcut}, while our dataset shows that human vision is also prone to errors due to shape bias; however, logical processing helps human vision recover from many of these errors. Deep learning models have outperformed human vision in many tasks that require the analysis of finer patterns \cite{shourie2023efficient,hannun2019cardiologist,ruamviboonsuk2019deep,stephansen2018neural}, and performance on these tasks can be improved with scaling \cite{haque2023effect} (for example, the biggest improvements have been made in figure ground, camouflage and similarity which require identifying finer patterns Fig.\ref{fig:prop-comparison}). The coarse to fine processing and early stopping can enable efficient processing in human vision. From an energy standpoint, the human brain is the most expensive organ \cite{kreiman2020beyond}; hence, visual processing must operate within energy and size constraints.

%% file: sec/6_llms.tex
\begin{figure*}[ht]
\centering
    \subfloat[Fraction of correct inference for segmentation models (e.g. second row in fig.\ref{fig:human-logical-cat-fg},\ref{fig:human-logical-tree-fg} are correct).]{\includegraphics[width =0.4\linewidth]{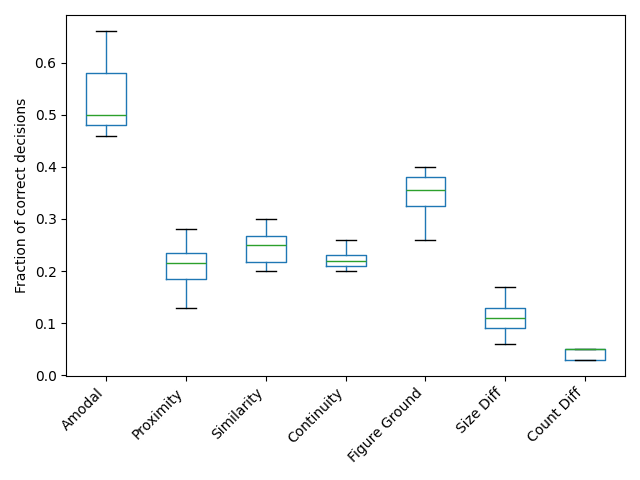}  \label{fig:box-correct}}
    \hfill
    \subfloat[Fraction of inferences similar to errors in human intuitive processing for segmentation models (e.g. third row in fig.\ref{fig:human-logical-cat-fg},\ref{fig:human-logical-tree-fg}).]{\includegraphics[width = 0.4\linewidth]{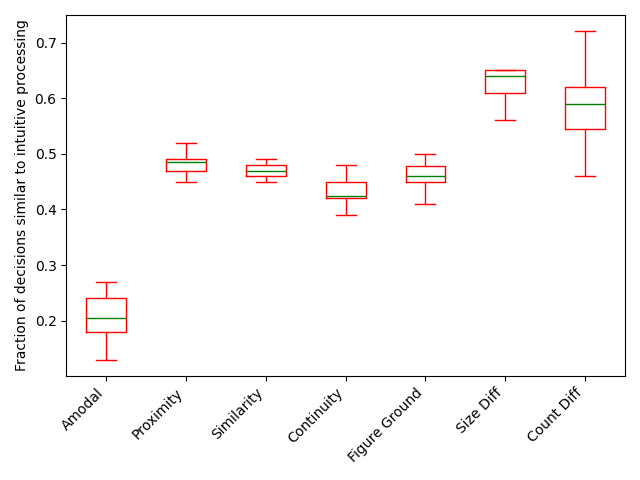} \label{fig:box-human} }    
    \hfill
    \subfloat[Plots comparing model performance qualitatively]{\includegraphics[width=0.9\linewidth]{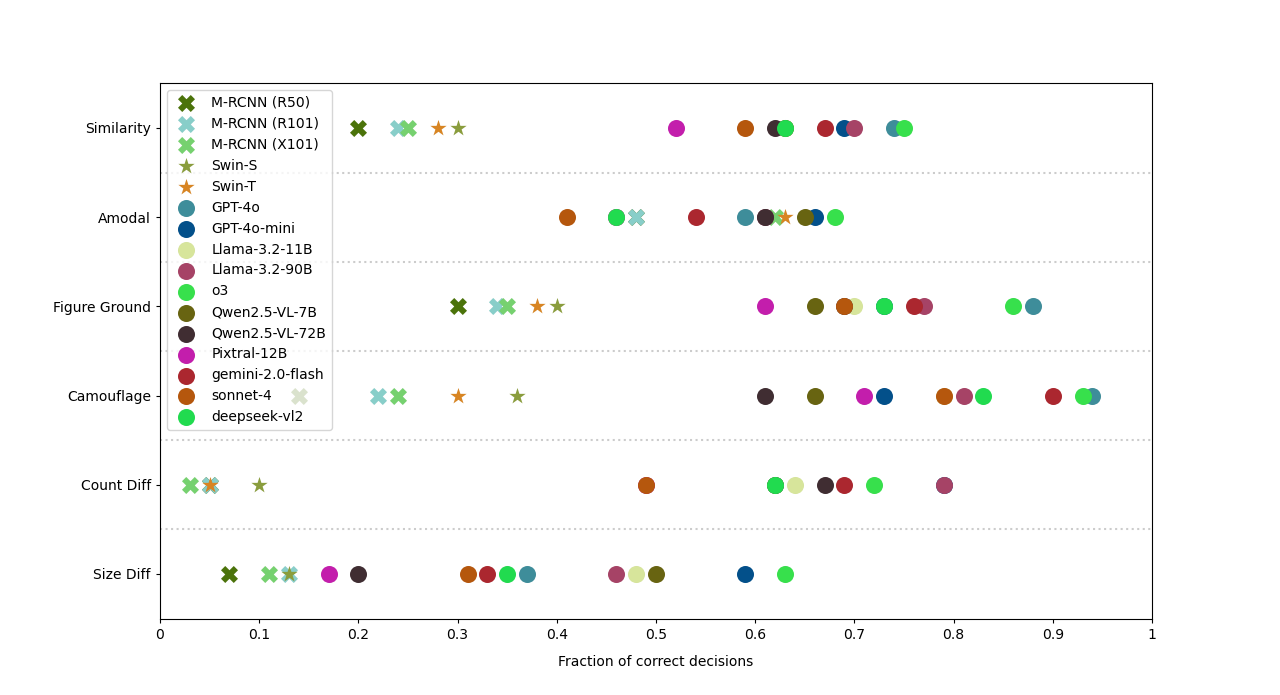} \label{fig:prop-comparison} }            

\caption{Multi-modal LLM's have made tremendous progress compared to segmentation models.}
\label{fig:modelbar}
\end{figure*}

\begin{figure*}[ht]
\centering
    
    \includegraphics[width = \linewidth]{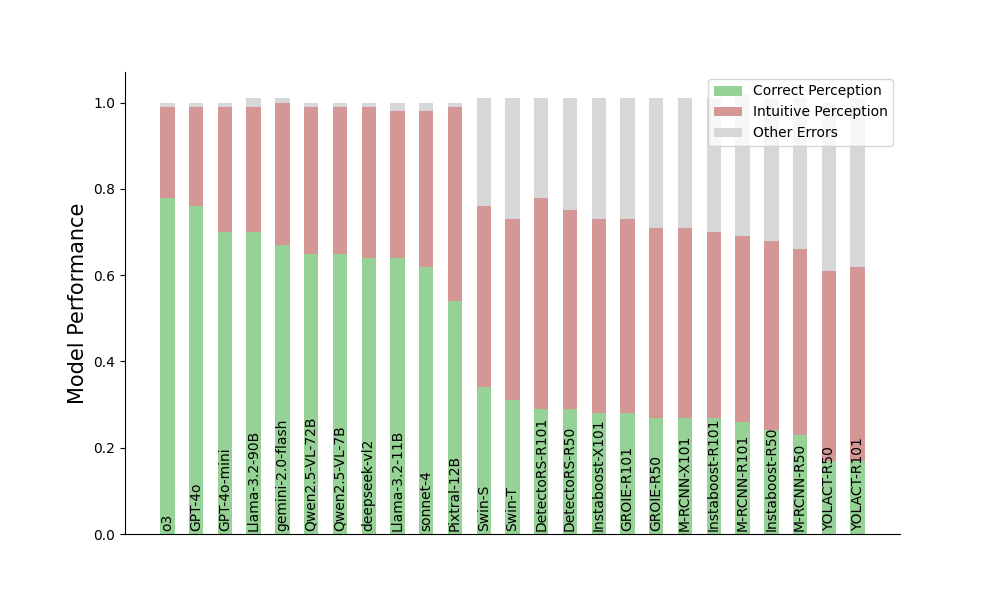}
\caption{Overall model performance. In segmentation models, the model can predict both human and correct and hence they can add up to more than 1.}
\label{fig:final-stacked}
\end{figure*}

%% file: sec/7_discussion.tex
\section{Discussion}

Initial approaches to reasoning in LLM involve planning the solution approach and guiding the answer generation through multiple steps or 'thoughts' (e.g. chain of thought). This approach was further improved by generating many answers and choosing the final response by voting (e.g. Self-Consistency with Chain of Thought \cite{wangself}). However, this method does not necessarily select the most suitable answer through reasoning and encounters performance saturation by scaling \cite{brown2024large}. In contrast, human visual processing makes an initial inference and then verifies for logical errors. In case of error, the solution search is constrained to the region with error to select the appropriate solutions for that specific region, which can also be more efficient. This is conceptually similar to the tree of thought method of reasoning \cite{long2023large}, which enables backtracking in case of error in a multi-step approach (more similar to eq.\ref{eq:avglenimprove} rather than eq.\ref{eq:terms}). However, unlike the step-by-step evaluations of the tree of thought method to generate the solution \cite{libenchmarking}, human visual processing appears to create a complete initial inference before verifying for errors. 

In this study, we investigate the role of reasoning in low-complexity visual perception to enhance robustness and address rare errors often overlooked in supervised learning. A key aspect of efficient perception is determining when reasoning is necessary. Human vision achieves this by assessing the integrity of perceived objects using basic factual knowledge. Incorporating verification steps, such as follow-up questioning, to enforce factual consistency (e.g., Fig.\ref{fig:multimodal-chat-error-correct}) can help correct count-related errors in LLMs, which possess relevant knowledge but fail to apply it during inference. Large reasoning models (LRMs) underperform on simple tasks compared to LLMs, but outperform them on moderately complex ones \cite{illusion-of-thinking}. While complex tasks (e.g., Fig.\ref{fig:human-hard-vqa}) demand more advanced reasoning, it is crucial to determine when such reasoning is needed. Human vision avoids the overthinking problem by choosing the first (or most probable) correct answer without errors and stopping early \cite{illusion-of-thinking}.
Additionally, human visual reasoning is influenced by cognitive biases. For instance, subtle cues such as clothing style, fit, or skin texture may be associated with gender in human perception, even though such features may not be present in representations learned by deep models. These differences underscore the distinct nature of reasoning in human vision compared to that in large language models (LLMs). Bridging this gap may involve aligning model behavior more closely with human reasoning using methods such as Reinforcement Learning from Human Feedback (RLHF), which can help models better handle ambiguous or visually subtle scenarios.

Reasoning and evaluation play a crucial role in identifying and addressing rare errors, often termed the "Curse of Rarity"—anomalies that arise infrequently in real-world ("in-the-wild") deployments yet remain critical to resolve in safety-sensitive systems \cite{liu2024curse}. Traditional machine learning approaches, which rely on pattern recognition from training data, may struggle with such cases due to entropy \cite{hoffmann2022training} or the scarcity of these patterns in training datasets, making it difficult for models to handle these scenarios effectively. This challenge is particularly significant as autonomous systems are increasingly deployed in safety-critical domains such as autonomous driving. Our dataset provides a more accurate representation of real-world scenarios than specialized datasets for testing model capabilities without requiring domain-specific information or external tool integration. For instance, solving questions in the MMMU dataset for multimodal LLMs \cite{yue2024mmmu} (Fig.\ref{fig:human-hard-vqa}) often necessitates a structured, step-by-step reasoning approach and specialized domain knowledge. In contrast, our dataset primarily relies on immediate perception, which aligns more closely with real-world visual cognition.

\begin{figure}[ht]
    \centering
        \subfloat[\label{fig:park-assist-error}]{\includegraphics[height=1.8cm]{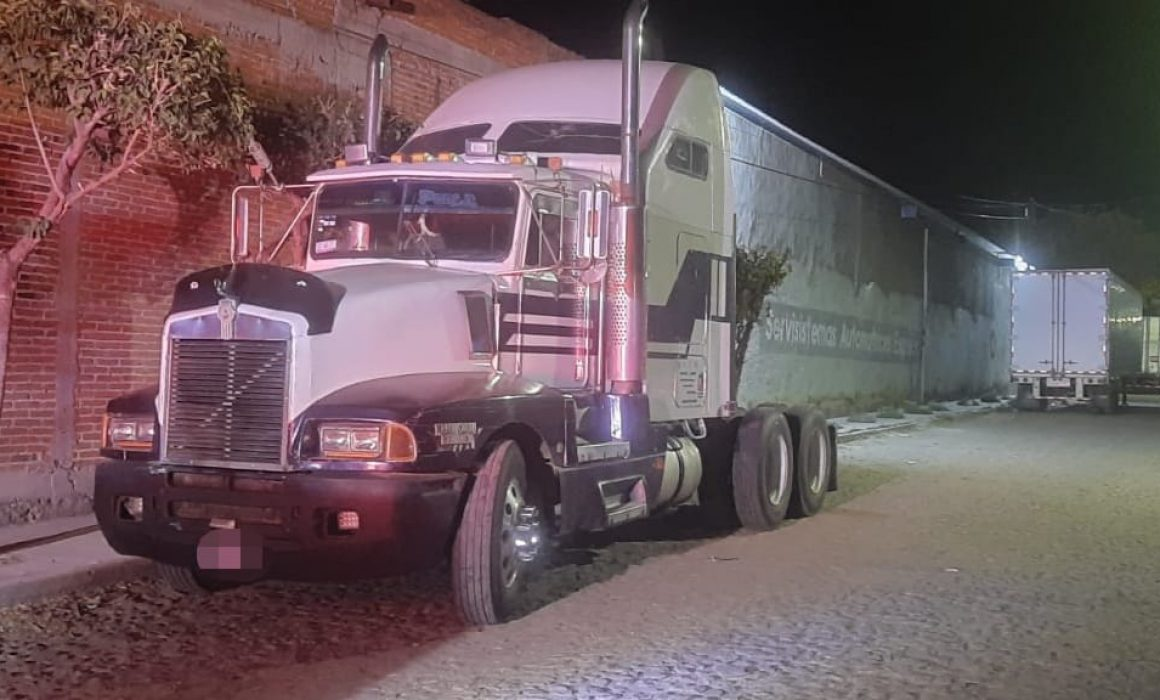}} \hfill 
        \subfloat[\label{fig:autonomous-driving-error}]{\includegraphics[height=1.8cm]{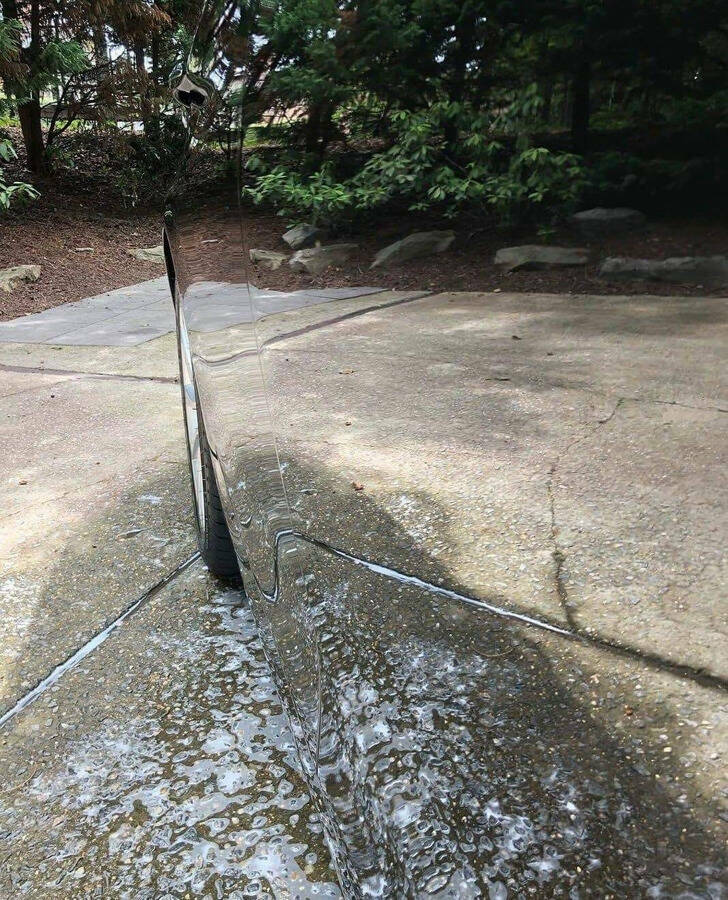}} \hfill 
        \subfloat[\label{fig:face-spoofing-error}]{\includegraphics[height=1.8cm]{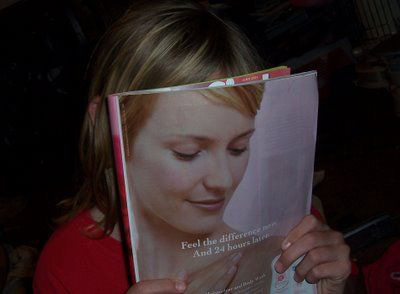}} \hfill 
        \subfloat[\label{fig:retail-counting-error}]{\includegraphics[height=1.8cm]{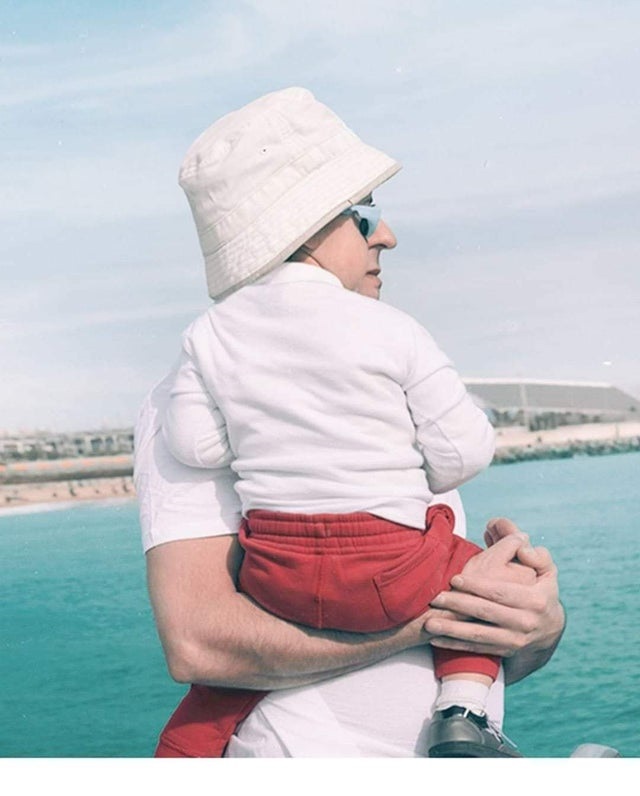}} \hfill 
    \caption{
        Examples from the human confusion dataset that can be useful for specific domains.
        (a) Automatic parking assist can ignore available spaces.
        (b) Driving assist can fail to recognize presence of cars and can cause collision.        
        (c) Facial spoofing can compromise facial recognition systems using aligned photos or masks.
        (d) Person recognition systems can fail to identify instances correctly.
        }
    \label{fig:VisualTheories}
\end{figure}

Large language models (LLMs) are computationally demanding and energy-intensive, making them impractical for many real-time or resource-constrained environments. In such cases, smaller, task-specific models are more suitable—such as those used in autonomous driving, parking assistance, facial recognition, robotics, retail surveillance, and drone operations. These systems are more likely to encounter visually ambiguous or partially obscured objects in real-world settings, increasing the risk of misinterpretation. For instance, autonomous vehicles may misidentify static objects or other vehicles as drivable paths (e.g Fig.\ref{fig:autonomous-driving-error}) or drivable areas for parking vehicles (e.g Fig.\ref{fig:park-assist-error}). Security and surveillance systems may fail to detect individuals due to facial spoofing techniques, such as printed photos, background blending, clothing-based obfuscation, or challenging lighting. In high-stakes settings like airports, facial recognition systems can be tricked by fake or printed faces (e.g Fig.\ref{fig:face-spoofing-error}), resulting in false identifications or undetected access breaches. These examples highlight the need for robust and reliable models capable of handling real-world ambiguity, particularly where deploying large models is infeasible. Systems could also be designed to flag potentially confusing inputs for human review or more intensive processing, adding a layer of safety and effective design of human-in-the-loop systems. Our dataset provides a valuable source for assessing such edge cases, which can be extended and tailored to these specific safety-critical domains to support the development and evaluation of models.

\textbf{Limitations : } Our dataset is adversarial, requiring models to rely on cues that are not commonly encountered. As a result, strong performance on this dataset may not directly translate to improved metrics on standard benchmark datasets. While our study highlights the shortcomings of deep learning models compared to human visual processing, the solutions to these challenges may differ due to the distinct constraints within which each system operates. An optimal approach for deep learning models may not necessarily align with the mechanisms of human vision. For instance, human vision prioritizes rapid initial responses, which can be evolutionarily advantageous—for example, immediately recognizing a potential threat, such as a dangerous animal, and preparing to act. Addressing the limitations of deep learning may require specific solutions tailored to the constraints of the environment in which they operate. Developing robust deep learning systems for safety-critical and resource-constrained environments may require domain-specific extensions of the dataset, incorporating adversarial examples that reflect the unique challenges and failure modes relevant to the target application domain.

%% file: sec/8_conclusion.tex
\section{Conclusion}
\label{sec:conclusion}

Our adversarial dataset provides insights into the dual-thinking framework in human visual processing, and computational analysis shows the advantages of human visual processing. The analysis of model behavior shows the limitations of segmentation models in performing logical processing in the second stage of human vision and the tremendous progress made by multi-modal LLMs. However, LLMs still lack fine-grained reasoning abilities, including recognizing size differences. Our study shows that segmentation models can replicate characteristics of human vision, indicating their potential to serve as computational models for human vision. Additionally, our dataset can help cognitive scientists evaluate new theories on human vision and validate assumptions about human visual processing.
